\documentclass[preprint,12pt]{elsarticle}

\usepackage{amsmath,amssymb,amsfonts}
\usepackage{algorithm}
\usepackage{algorithmicx}
\usepackage[noend]{algpseudocode}
\usepackage[caption=false]{subfig}

\usepackage{graphicx}
\usepackage{textcomp}
\usepackage{xcolor}
\usepackage{tcolorbox}
\usepackage{graphicx}

\usepackage{hyperref}

\usepackage{comment}

\usepackage{latexsym}
\usepackage{subfig}
\usepackage{float}

\usepackage{colortbl}
\usepackage{multirow}
\usepackage{adjustbox}

\usepackage{chngcntr}
\usepackage{booktabs}
\usepackage{enumitem}
\usepackage{xurl}
\usepackage{pgfplots}
\pgfplotsset{compat=1.18}
\usepackage{subcaption}

\usepackage{tabularx, rotating}
\algrenewcommand{\algorithmiccomment}[1]{\hskip1em$\triangleright$ \textcolor{gray}{#1}}

\graphicspath{{figs/}}

\newtheorem{definition}{Definition}

\journal{Information Systems}

\begin{document}

\begin{frontmatter}

\title{Business Process Simulation: Probabilistic Modeling of Intermittent Resource Availability and Multitasking Behavior}

\author[Estonia]{Orlenys L\'opez-Pintado\corref{corresp}}
\cortext[corresp]{Corresponding author}
\ead{orlenyslp@ut.ee}

\author[Estonia]{Marlon Dumas}

\address[Estonia]{University of Tartu, Estonia}

\graphicspath{{figs/}}

\begin{abstract}
In business process simulation, resource availability is typically modeled by assigning a calendar to each resource, e.g., Monday-Friday, 9:00-18:00. Resources are assumed to be always available during each time slot in their availability calendar. This assumption often becomes invalid due to interruptions, breaks, or time-sharing across processes. In other words, existing approaches fail to capture intermittent availability. Another limitation of existing approaches is that they either do not consider multitasking behavior, or if they do, they assume that resources always multitask (up to a maximum capacity) whenever available. However, studies have shown that the multitasking patterns vary across days. 
This paper introduces a probabilistic approach to model resource availability and multitasking behavior for business process simulation. 
In this approach, each time slot in a resource calendar has an associated availability probability and a multitasking probability per multitasking level.
For example, a resource may be available on Fridays between 14:00-15:00 with 90\% probability, and given that they are performing one task during this slot, they may take on a second concurrent task with 60\% probability. We propose algorithms to discover probabilistic calendars and probabilistic multitasking capacities from event logs. An evaluation shows that, with these enhancements, simulation models discovered from event logs better replicate the distribution of activities and cycle times, relative to approaches with crisp calendars and monotasking assumptions.

\end{abstract}



\end{frontmatter}

\section{Introduction}

Business Process (BP) simulation is a technique to estimate the impact of changes to a process on one or more performance metrics. It enables analysts to answer ``what-if'' questions such as ``What would be the cycle time of a process if some of the resources switched from full-time to part-time?'' or ``What would be the cycle time if the resources responsible for a key activity in the process would handle up to three activity instances simultaneously instead of focusing on one activity instance at a time, and in doing so, they increased their throughput from 6 to 9 activity instances per hour?

The starting point of a BP simulation is a process model, represented, for example, in the Business Process Model and Notation (BPMN)~\cite{bpmnspec13}, enhanced with simulation parameters such as the processing times of each activity, the inter-arrival time between cases, the available resources and their availability~\cite{Aalst15}. Each execution of a BP simulation model (a.k.a., a simulation run) produces an event log recording the simulated execution of several process cases alongside aggregate performance statistics, such as the mean cycle time of the simulated cases.

In mainstream BP simulation approaches, including the BPSim standard~\cite{BPSim}, resource availability is captured by assigning a calendar to each resource or group of resources, e.g., from Monday to Friday, 9:00-18:00. In this approach, a resource is available during every time slot in their calendar and not available at all outside their calendar. This boolean ``on-or-off'' approach fails to capture the intermittent nature of resource availability in real-world scenarios, where resources are affected by interruptions, breaks, meetings, or time-sharing across multiple processes.

Similarly, the bulk of existing BP simulation approaches, including BPSim, overlook the possibility that a single resource may perform multiple activity instances concurrently (multitasking). For example, in BPSim, one may specify that there are five resources of role ``Clerk''. In a state of a simulation run where all five clerks are busy, any new activity instance that requires a clerk is put on hold until one of the clerks becomes available. In other words, the modeler cannot specify that a single clerk may perform two or more activity instances concurrently. Yet, numerous studies in the field of organizational behavior have shown that multitasking is a common behavior in office work~\cite{Spink2008}.
Some BP simulation approaches capture multitasking by assigning a \emph{multitasking capacity} to each role~\cite{BocciarelliDW22}. These approaches assume that resources will always take on multiple activities concurrently, up to their multitasking capacity. In the above example, if all clerks are busy and some activity instances are waiting to be started, the already busy clerks will continue taking on more activity instances until they reach their multitasking capacity.\footnote{The concept of multitasking capacity can be emulated in BPSim by specifying that the number of resources of a role is equal to the actual number of resources times their multitasking capacity. For example, to emulate a situation where there are five clerks, each of whom may perform up to two activity instances concurrently, we can set the number of resources of role Clerk to be equal to $5 \times 2 = 10$.} 
However, empirical studies have shown that the multitasking behavior of workers varies over time~\cite{KirchbergRE15}. A worker who normally takes on multiple activity instances concurrently on the first day of their work week might not do so towards the end of the week.




To tackle the above limitations of existing BP simulation approaches, this article proposes: {\bf 1}) a business process simulation approach wherein resource availability is captured via probabilistic calendars instead of crisp ``on-or-off'' calendars; {\bf 2}) a method to discover such probabilistic availability calendars from event logs; {\bf 3}) a probabilistic simulation approach to capture probabilistic multitasking capacity; and {\bf 4}) a method to discover probabilistic multitasking capacities from event logs.

In the proposed approach, an availability calendar assigns a probability to each time slot in a calendar, e.g., a resource may be available with 90\% probability on Fridays from 9:00-10:00, and this probability decreases linearly every hour down to 30\% from 17:00-18:00 and then 5\% from Friday 18:00 to Monday 8:00. At each time point, the resource may be available or not, according to the probability of the corresponding time slot. 

Similarly, the proposed approach captures the multitasking capacity of a resource probabilistically by adopting the principle that resource availability and multitasking capacity are special cases of the same concept. Whereas resource availability refers to the probability of a resource taking on an activity instance given that they are not busy, multitasking capacity refers to the probability of a resource taking a second activity instance given that they already are performing one activity instance, or three activity instances given that they are performing two, etc.
For example, in the proposed approach, we can specify that a resource may perform up to three tasks simultaneously, with degressive probabilities, e.g., there is a 70\% probability that a resource can take on a second activity instance given that they are already performing one activity instance, and a 30\% probability of taking on a third activity instance given that they are already performing two activity instances. Furthermore, the approach enables us to specify that the multitasking capacity of a resource varies across the hours of the day or the days of the week, such that there is a 70\% probability of a resource taking on two activity instances simultaneously during busy morning hours but only a 40\% probability of doing so during the afternoon periods. 


The article reports on an empirical evaluation aimed at verifying two hypotheses: 1) that simulation models with probabilistic calendars derived from event logs have higher temporal accuracy than models with crisp calendars; 2) that models incorporating probabilistic multitasking have higher temporal accuracy than those that assume that workers always perform one activity instance at a time.

This article is an extended version of a conference paper~\cite{LopezPintadoD23}. The conference paper focuses on discovering and simulating business processes with probabilistic resource availability calendars. This article extends the conference paper with an approach to discovering and simulating business processes with probabilistic multitasking capacity.

The article is structured as follows. Section~\ref{sect:related} discusses related work. Section~\ref{sect:model} describes and formalizes a simulation meta-model that captures resource availability and multitasking behavior probabilistically. Section~\ref{sect:discovery} proposes methods to discover simulation models with probabilistic availability and multitasking calendars. Section~\ref{sect:evaluation} empirically compares crisp vs. probabilistic simulation models and multitasking vs. single-tasking execution models. Finally, Section~\ref{sect:conclusion} concludes and sketches future work.

\section{Related Work}
\label{sect:related}

Van der Aalst et al.~\cite{Nakatumba10, Aalst15} discuss limitations of existing BP simulation approaches, including (i) insufficient use of execution data for constructing simulation models; and (ii) incorrect modeling of resources. The authors observe that existing simulation approaches assume that each resource is entirely dedicated to the simulated process. At the time of the writing of this article, this assumption is embodied in several commercial BP simulation tools, such as ARIS\footnote{\url{https://aris.com}}, Apromore\footnote{\url{https://apromore.com}}, Bizagi\footnote{\url{https://www.bizagi.com}}, and iGrafx\footnote{\url{https://www.igrafx.com}}, as well as in the BPSim standard mentioned earlier.
In practice, though, a resource often time-shares across multiple processes or otherwise performs tasks that are not part of the simulated process. Hence, the resource will not always take on an activity instance in a process, because they are busy elsewhere, or they are taking an unplanned break. This paper addresses this limitation by proposing a BP simulation approach that uses probabilistic calendars to model resources that are intermittently available to perform activity instances in a process during their availability calendar, as well as an approach to discover such probabilistic calendars from execution data.


Arena~\cite{ArenaBook} -- a well-known discrete-event simulation tool --- captures intermittent resource availability via resource failure models. In this approach, the availability of a resource is dependent on a parameter, whose value may change over time, specifically when the resource moves between an operating state and a failure state. To capture intermittent downtimes, the modeler has to define a failure model, which may, for example, update the availability of a resource at random time intervals.
This article adopts a more direct approach in which resource availability is always probabilistic, independent of any failures. This approach draws inspiration from Lee et al.~\cite{LeeL04} who define a notion of fuzzy calendars as functions that assign a probability to each interval in a set of intervals (e.g., every Monday at 9-10 am during the year 2023). In this line, our approach defines a probabilistic calendar as a mapping from (periodic) sets of intervals to probabilities, and uses such calendars to capture intermittent resource availability in a BP simulation approach.

Another limitation of BPSim and other mainstream business process simulation approaches, is that they either assume that resources do not multitask, or, if they multi-task, they assume that their multitasking capacity does not change over time. Arena departs from this assumption by modeling multitasking capacity as a resource parameter (namely MR) whose value may change over time, either based on a schedule or by capturing a change in capacity when a failure occurs (e.g., one can model that the capacity of a resource is reduced by 50\% when a failure occurs). However, at any given point in time, the capacity of a resource has an integer value MR, with the semantics that the resource will take up to MR jobs (activity instances). In this article, we take inspiration of this ``variable multitasking capacity'' approach, but we model the multitasking capacity probabilistically, meaning that a resource may, but will not necessarily take on additional tasks, even if they have spare multitasking capacity. 



With respect to the above approaches, another contribution of this article is that we discover the availability and multitasking capacities from execution data, meaning that our proposal encompasses not only the simulation component, but also a simulation model discovery component. In this respect, this article is related to prior studies that propose methods to discover BP simulation models from event logs. The first generation of such studies~\cite{RozinatMSA09, MartinDC16, CamargoDG20} assumed that resources are continuously available (24/7). More recent studies~\cite{MartinDCS20, Estrada-TorresC21} incorporate algorithms for discovering resource calendars from event logs. However, the above approaches discover ``crisp'' availability calendars, i.e., a resource is either available or not during each time slot in the calendar. Also, these approaches do not account for multitasking behavior -- except for~\cite{Estrada-TorresC21} as discussed below.

Recent studies acknowledge the above limitations in BP simulation models and propose to address them by modeling the waiting times and processing times of activities via a deep learning (DL) model~\cite{CamargoBDR23,MeneghelloFG23}. While these approaches indeed have the potential to capture both intermittent availability and multitasking (i.e., variable resource capacity), they suffer from the black-box nature of DL models, which makes it impractical for end-users to alter these BP simulation models to capture ``what-if'' scenarios. 




Freitas \& Pereira highlight another common limitation of BP simulation modeling approaches: the assumption that all resources in a group (a.k.a., a resource pool) have the same availability calendar~\cite{Freitas15}. Some tools like IBM Websphere Modeler\footnote{\url{https://www.ibm.com/support/pages/download-websphere-business-modeler-advanced-v70}} support named resources, each having its own availability calendar. In this paper, we adopt this latter approach (calendars may be attached to each resource individually), but we extend it to support probabilistic availability calendars. Specifically, we take as starting point an approach we proposed earlier~\cite{Lopez-PintadoD22}, wherein each resource has its own availability calendar and its own performance profile. This approach is implemented in the Prosimos business process simulator~\cite{Lopez-PintadoHD22}. 

Nakatumba et al.~\cite{NakatumbaWA12} note that, according to the Yerkes-Dodson Law of Arousal, people take more time to execute an activity if there is little work to do. Accordingly, the authors propose a simulation modeling approach wherein the processing time of an activity depends on the current workload. The authors also propose an approach to learn the relation between workload and processing times from event logs. In this article, we address a complementary problem, namely that of determining the availability of resources. Our approach also captures relations between time slots in a calendar and processing times (the processing time may depend on the time of the day) but we do not capture relations between workload and processing times.


Rusinaite et al.~\cite{RusinaiteVSVN16} discuss how resources can be ``shared'' by multiple tasks across multiple processes, similar to our notion of multitasking, where a single resource can handle multiple activity instances within a given process. However, their approach does not capture variable multitasking capacity nor does it discover the multitasking capacity from execution data.

Ouyang et al.~\cite{OuyangTFHK10} focus on shared material resources in multitasking situations, such as surgical materials used by multiple doctors in an operation. Unlike our method, they do not explore the analysis of real execution data to discover simulation models with multitasking behaviors.

Estrada-Torres et al.~\cite{Estrada-TorresC21} addresses the problem of discovering BP simulation models from event logs, in the presence of multitasking. Their approach preprocesses the event log to identify situations where a resource performs multiple activity instances concurrently. When this happens, their approach shortens the processing times of the multitasked activity instances, so that, if the shortened activity instances are executed sequentially, their total processing time is equal to the total time it took to perform these same activity instances with multitasking. For example, if a resource performs an activity instance in the interval [5, 8] and another in the interval [6,10], the approach of Estrada-Torres et al. will consider that the first activity instance took 2 time units, and the second one 3 time units, so that collectively, the two activity instances take 5 time units (the difference between 10 and 5). More generally, this preprocessing step produces a log without multitasking, such that for each resource, the total amount of time during which this resource is busy is the same as in the original log.
After this preprocessing, the resulting event log is given as input to the Simod tool~\cite{CamargoDG20}, which discovers a simulation model assuming no multitasking.
The authors show that this approach leads to simulation models that generate cycle time distributions similar to those of the original event log. However, the simulation does not consider multitasking, and hence, the user cannot use the discover simulation model to analyze what-if scenarios such as ``what would happen if we increase the multitasking capacity of some of the resources in the process?''. In contrast, our approach integrates multitasking directly into the discovered simulation model and captures scenarios where the multitasking capacity varies over time.

\section{Simulation with Probabilistic Calendars and Multitasking}
\label{sect:model}

We take as a starting point a resource model wherein each resource has a \emph{resource profile}. The profile of a resource determines the activities that the resource may perform, the performance of this resource (i.e., how much time the resource takes to perform different activities), the cost per time unit of the resource, and the availability calendar of a resource. 
This model treats resources in a differentiated manner, insofar as each resource has its own profile, but it does not prevent multiple resources to have identical profiles.
An activity may be performed by more than one resource, and multiple activities may share common resources.
The time that a resource takes to perform an activity is captured as a number drawn from a distribution.
Availability calendars consist of the time intervals during which a resource is available, under the assumption that a resource is either available continuously or not available during each of these intervals (i.e., crisp calendar). These considerations are formalized below.


\begin{definition}[BP Simulation Model - adapted from~\cite{Lopez-PintadoD22}]\label{def:diff}
A BP simulation model with differentiated resources $DSM$ is a tuple $<E, A, G, F,$ {\sc RProf}$, BP, AT, AC>$, where $E, A, G$ are the sets of events, activities, and gateways of a BPMN model, $F$ is the set of directed flow arcs of a BPMN model, and the remaining elements capture simulation parameters as follows:   
 \begin{enumerate}
     \item {\sc RProf} $: \{r_1, ..., r_n \}$ is a set of resource profiles, where $n$ is the number of resources in the process. Each resource $r \in RProf$ is described by:
     \begin{itemize}
         \item {\sc Alloc}$(r) \subseteq A$ is the set of activities that $r$ can execute 
         \item {\sc Perf}$(r, \alpha): $ {\sc Alloc}$(r)$ $\rightarrow PDF(\mathbb{R}^+)$ is a function that maps each activity $\alpha \in$ {\sc Alloc}$(r)$ to a probability density function $\mathcal{P} \in PDF(\mathbb{R}^+)$. This $PDF$ represents the distribution of the processing times of activity $\alpha$ when assigned to $r$, with the processing time being a continuous random variable.
          \item {\sc Avail}($r$)$: Time \rightarrow [0, 1]$ is a function that maps each time slot in a calendar to a value indicating the availability of resource $r$ to perform activities in {\sc Alloc}$(r)$. A value of 1 indicates definite availability, 0 indicates no availability, and any value between 0 and 1 indicates availability with a certain probability.         
         \item {\sc Multi}($r$): $Time \rightarrow \mathbb{N^+} \cup PDF(\mathbb{N^+})$  is a function mapping each time slot to a value indicating the multitasking capacity of the resource $r$. It can be a fixed number denoting the maximum number of simultaneous activities $r$ can perform or a probability density function indicating the distribution of multitasking capacities.
         
         
         \item {\sc Cost}($r$): $\mathbb{R}^+$ is the unit cost (e.g., per hour) of resource $r$ 
     \end{itemize}
          
    \item {\sc BP}$ : F \rightarrow [0, 1]$ is a function that maps each flow $f \in F$ s.t., the source of $f$ is an element of $G$ to a probability (a.k.a., the branching probability).
     
     \item {\sc AT}: $PDF(\mathbb{R}^+)$ is a probability density function modeling the inter-arrival times between consecutive case creations. 
     
     \item {\sc AC} is a calendar (set of intervals) such that cases can only be created during an interval in {\sc AC}.
 \end{enumerate}
\end{definition}

To illustrate the elements of a resource profile in Definition~\ref{def:diff}, consider a customer service process. Here,  {\sc RProf} contains, among others, the resource profile of an agent $r$ described by  {\sc Alloc}$(r) =\{LogRequest\}$, indicating that the agent can execute only the logging activity. The performance function {\sc Perf}$(r, LogRequest) = \mathcal{N}(20, 5^2)$ maps the activity ``LogRequest'' to a normal distribution with mean $\mu = 20$ and standard deviation $\sigma = 5$, i.e., meaning most of the activities performed by the agent will take around 20 time units, with a decreasing likelihood for times further from this mean. The availability {\sc Avail}$(r) =$ {\tt <9:00-17:00, Monday-to-Friday>} indicates that, according to some scheduling policy (see section~\ref{sect:prob-calendar}), the agent is available during these times, where they can handle up to 3 activities simultaneously by defining {\sc Multi}$(r) = 3$ (see Section~\ref{sect:prob-multi}). Finally, {\sc Cost}$(r) = 20$ dollars per hour establishes the agent's cost.

A Business Process (BP) simulation model, as per Definition~\ref{def:diff}, produces an event log when executed by a simulation engine. An event log consists of a set of events. Each event represents an activity instance, i.e., an execution of an activity in the context of a case. Each event contains an identifier of the case where the activity instance occurs, an activity label, the resource that performed the activity instance, the start datetime of the activity instance, and its end datetime\footnote{We use the term \emph{timestamp} to refer to a time relative to a start of a day, e.g., 13:00:00 and \emph{datetime} to refer to a time including date and time of day.} indicating when the activity was started and ended. A trace is the set of events in a log that share the same case identifier. An event log is a set of traces, each representing a process instance (case). We write \emph{simulated log} to refer to a log produced from a simulation model and \emph{real log} to refer to a log extracted from an information system.

We require each activity instance in an event log to have start and end times because we must reason about activity instance processing times when performing a simulation. While some systems, like Enterprise Resource Planning systems (ERP), may only record end times, others, such as case management and workflow management systems, typically log both. Additionally, techniques exist in the literature to estimate start times when they are absent~\cite{FraccaLAT22}. Our approach also considers enabling times, which are not required in the event log, as we estimate them by discovering a process model and replying to the event log over it.

Several performance metrics can be derived from event logs, for example, \textit{waiting time} -- the time-span from enabling time to the start of an event; \textit{processing time} -- the time-span between beginning and end of an event; \textit{cycle time} -- the time-span between the enabling time and end of an event;\footnote{The concept of cycle time can also be applied to a \emph{case}. The cycle time of a case is the difference between the largest end datetime and the smallest start datetime in the case. By extension, the cycle time of a process is the mean of the cycle time of its cases.} and \textit{resource utilization} -- the ratio between the time a resource is busy executing activity instances, divided by its total availability time.

This paper focuses on discovering and representing the functions {\sc Avail}, {\sc Perf}, and {\sc Multi} in Definition~\ref{def:diff}. Accordingly, Definitions~\ref{def:granularity},~\ref{def:prob_calendar},~\ref{def:res_availability} formalizes the concepts of time granularity, probabilistic granularity, and probabilistic resource calendar proposed in this paper to model {\sc Avail}. The {\sc Perf} function is modeled by {\sc adjustProcessingTime} in Definition~\ref{def:res_availability}. Finally, Definitions~\ref{def:multi_prob} and~\ref{def:res_multi_capacities} capture the dynamics of the resource multitasking capacity related to {\sc Multi} using discrete probabilistic distributions.

\subsection{Probabilistic Resource Availability Calendar Model}
\label{sect:prob-calendar}

Time Granularity (Definition~\ref{def:granularity}) refers to segmenting time into defined, discrete intervals, often called `granules'. Each granule represents an interval defined by two timestamps within a 24-hour period (e.g., from 13:00:00 to 14:00:00), and all granules within this period have the same duration (which can be interpreted in a time unit, such as seconds, minutes, or hours). These granules do not overlap or intersect and provide the minimum interval for positioning the timestamp of each event executed in a process.


\begin{definition}[Time Granularity]\label{def:granularity}
A time granularity $\Delta_n$ is a sequence of equidistant consecutive timestamps $\tau_0, \tau_1, ..., \tau_{n}$, where $n$ (a.k.a., the granularity size) is the total number of time granules. Each pair of consecutive timestamps $\tau_i$ and $\tau_{i+1}$ defines a time granule $\delta_i = [\tau_i, \tau_{i+1})$ with a duration $d$. The following rules must hold: the total duration of the granularity is 24 hours, i.e., $|\tau_{n} - \tau_0| = 24$ hours, and for each $\tau_i \in \Delta$ with $i > 0$, $\tau_i = \tau_0 + (i * d)$, ensuring that $\tau_i - \tau_{i-1} = d$.
\end{definition}


It holds from Definition~\ref{def:granularity} that a granularity $\Delta$ is defined within the context of a single day. For example, it can represent that a resource works from 8:00 to 17:00, so in this context, whether it is Monday or any specific day of the year is irrelevant. However, we must also consider repeating time intervals across multiple days, weeks, months, or years to model calendars in real-world processes. To this end, Definition~\ref{def:prob_calendar} introduces the concept of recurring slots $\Omega$, which allows the definition of intervals that repeat over specified periods, extending the applicability of granules beyond a single day.


\begin{definition}[Probabilistic Granularity]\label{def:prob_calendar}
Let $\Omega$ be a set of recurring slots and $\Delta$ a time granularity. A probabilistic granularity is a function $P: \Omega \times \Delta \to [0, 1]$ mapping each pair $(\omega \in \Omega, \delta \in \Delta)$, named p-granule, to a real number in $[0, 1]$ representing a probability. Recurring slots $\omega \in \Omega$ define a time interval with a duration greater than or equal to the size of the granularity $\Delta$. Besides, $\forall$ $\omega_1, \omega_2 \in \Omega$, $\omega_1 \cap \omega_2 = \emptyset$.
\end{definition}

Combining $\Delta$  and $\Omega$ creates a flexible schema to model calendars. On the one hand, the granularity $\Delta$ provides a time segmentation within a day. On the other hand, recurring slots $\Omega$ contextualize these intervals within broader periodic patterns. For example, assume the days of the week as recurring slots, $\Omega_w = \{${\tt <Monday>, <Tuesday>, ..., <Sunday>}$\}$. Let $\Delta = \{${\tt [00:00-01:00), [01:00-02:00), ..., [23:00-24:00)}$\}$ be a granularity with 1-hour granules. The p-granularity $P(${\tt Monday, 09:00-10:00}$) = 0.5$ sets an availability probability of 0.5 every Monday from 9:00 to 10:00. Consider now quarterly recurring slots $\Omega_Q = \{$ {\tt <Q1: January-March>, ..., <Q4: October-December>}$\}$, here $P(${\tt Q1: 09:00-10:00}$) = 0.5$ means that every single day in the first quarter of a year, the availability probability is 0.5 from 9:00 to 10:00, so the days of the week are irrelevant. Similarly, it can model different availability every single day of the year by defining $\Omega_D = \{${\tt <January, 01>, <January, 02> ..., <December, 31>}$\}$.

Recurring slots can also capture multiple levels of periodicity. For example, $\Omega = \{$ {\tt <January, Monday-Friday>, <January, Saturday-Sunday>, <February-June>, <July>, <August, 01>, ..., <December, 31>} $\}$ defines different time intervals through the year, such as weekdays and weekends in January, broader weekly calendar from February to June, potential vacations in July, and daily slots for the remaining months. These slots can even include multi-year periods to manage varying resource schedules for long-term projects that span several years. However, it is required to ensure that these intervals do not overlap. For example, defining both $\Omega = \{${\tt<Q1: January-March>, <January, Monday-Sunday>}$\}$ creates an invalid intersection, making it ambiguous to decide which pattern to apply. To avoid this, we can reorganize them into non-overlapping slots such as $\Omega = \{${\tt <January, Monday-Sunday>,  <February-March>}$\}$, which, when unfolded, lead to disjoint time intervals.


In the following, Definition~\ref{def:res_availability} combines the ideas of time granularity, probabilistic granularity, and methods to model resource availability by probabilistic calendars.

\begin{definition}[Probabilistic Resource Calendar]\label{def:res_availability}
A probabilistic resource calendar consists of the following functions:
\begin{itemize}
    \item {\sc absoluteProbabilty}, $P_{ABS}$, is a probabilistic granularity that quantifies the probability of a resource being available in a p-granule, given that a task that can be allocated to them is enabled. 
    
    \item {\sc relativeProbabilty}, $P_{REL}$, is a probabilistic granularity that quantifies the probability of a resource to be available in a p-granule $<\omega, \delta>$ relative to how often other resources are available in $<\omega, \delta>$.
    
    \item $\Gamma$ is a function that retrieves the corresponding recurring slot and time granule from a given datetime.
    
    \item {\sc isAvailable} is a function that, given a p-granule $<\omega, \delta>$, applies Bernoulli distributions to decide whether the resource is available (or not): \\ ({\tt True} with $P_{ABS}(\omega, \delta)$, {\tt False} with 1 - $P_{ABS}(\omega, \delta)$) {\bf or} ({\tt True} with $P_{REL}(\omega, \delta)$, {\tt False} with 1 - $P_{REL}(\omega, \delta)$).

    \item {\sc nextAvailableTime} is a function that, given a datetime $\sigma$, retrieves the nearest datetime $\sigma'$ in which the resource will be available as follows: $\min \sigma' : \sigma' \geq \sigma$ $\land$ {\sc isAvailable}$(\Gamma(\sigma')) =$ {\tt True}.

    \item {\sc adjustProcessingTime} is a function that receives a datetime $\sigma$ and a floating number $pt$ representing an ideal processing time, i.e., assuming the resource is fully dedicated and available during that $pt$ period. Then, it returns a datetime $\sigma'$ after adjusting $pt$ by adding the time the resource is unavailable according to their calendar, i.e.,  $\sigma'$ = $\sigma + pt$ + $\sum_{\delta}^{{\sc isAvailable}(\delta) = False} |\delta|$, $\delta \in \{\Gamma(\sigma_{i+1}): \sigma_{i+1} = \sigma_i + d\}_{\sigma}^{\sigma'}$, being $d$ the time interval duration defined by the corresponding granularity.
\end{itemize}
\end{definition}

\begin{figure}[t]
 \centering
\centerline{\includegraphics[scale=.56]{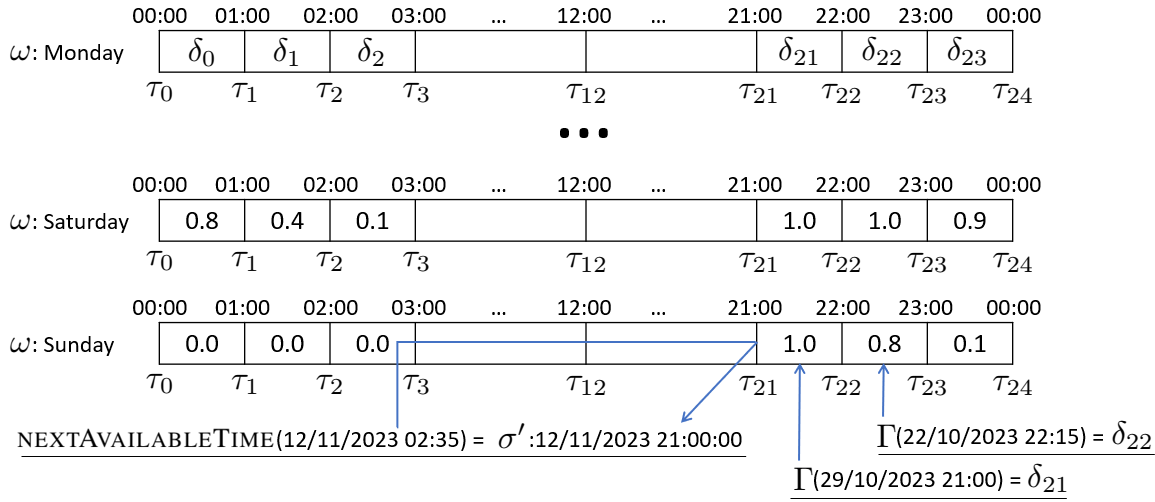}}
 \caption{Probabilistic Weekly Calendar of granularity $\Delta_{24}$, i.e., starting at $\tau_0 =$ {\tt 00:00:00}, with p-granules of size $d = 1$ hour.}
 \label{fig:granule-ex}
\end{figure}

Figure~\ref{fig:granule-ex} sketches the time granules of a probabilistic weekly calendar. In this calendar, the recurrent slots are the weekdays, i.e., they repeat every seven days, each split into n = 24 granules ($\delta_0, ..., \delta_{23}$), each of size d = 1 hour, and starting at midnight $\tau_0 = 00:00:00$. Granules on Saturday and Sunday have an associated probability value (p-granules). In this example, the resource is always available on Saturdays from 21:00 to 23:00 and never available on Sundays from 00:00 to 03:00. The bottom of Figure~\ref{fig:granule-ex} illustrates how function $\Gamma$ retrieves the corresponding p-granules given two datetimes, i.e., both correspond to a Sunday, but to the granules $\delta_{21}$ and $\delta_{22}$ in which the resource is available with a probability of 1.0 and 0.8 respectively. Finally, the function {\sc NextAvailableTime} probabilistically determines that if requested on 12/11/2023 at 02:35, the resource will be available the same day but at 21:00 at the earliest. Note that although not illustrated in the figure, function {\sc NextAvailableTime} may retrieve a different (future) date, i.e., one that does not correspond to the input datetime. 

Characterizing a resource calendar by combining absolute ($P_{ABS}$) and relative ($P_{REL}$) probabilities in Definition~\ref{def:res_availability} deserves further explanation. On the one hand, $P_{ABS}$ measures availability by counting the frequency ratio of a resource from all the occurrences of related events in a given p-granule. However, a uniform task allocation to resources may negatively impact this value. For example, assume a task always scheduled on Mondays from 9:00-10:00 and a pool of 10 resource candidates. In this case, if a different resource is appointed alternately every Monday, each will exhibit a low probability of 0.1 after the first rotation, even when they were always available. In cases like that, a relative value comparing the resources with the busiest one in the granule leads to a more accurate estimation. On the other hand, $P_{REL}$ evaluates resource frequency relative ratio to the most occupied resource in a granule, potentially disadvantaging resources with lighter workloads. For example, a frequently allocated resource within a given p-granule could substantially lower the probability of other resources that were consistently available but not needed due to the lack of enabled tasks. Consequently, our method combines absolute and relative probabilities to capture a more comprehensive range of resource availability.

\subsection{Probabilistic Resource Multitasking Capacity Model}
\label{sect:prob-multi}

While modeling resource availability determines `when' resources can execute process activities, it only partially captures their operational capacity. The next step is to model `how' these resources handle their workload during their available periods. To address this, we introduce the concepts of the Multitasking Discrete Probability Distribution (MDPD) and Resource Multitasking Capacities in Definitions~\ref{def:multi_prob} and~\ref{def:res_multi_capacities}.

\begin{definition}[Multitasking Discrete Probability Distribution]\label{def:multi_prob} 
 Let \\ $\mu_i$ be a discrete random variable representing the number of instances of activity $\mu$, and let $n$ represent the maximum number of activity instances that can be performed simultaneously (multitasking limit). The Multitasking Density Probability Distribution, denoted as {\sc MDPD}$_n(\mu_i)$, represents the probability of activity $\mu$ being executed concurrently $i$ times. The {\sc MDPD} is an inverse cumulative distribution function formally defined as: \[ MDPD_n(\mu_i) =
    \begin{cases}
        \frac{\sum_{j = i}^{n} \text{Freq}(\mu_j)}{\sum_{j = 1}^{n} \text{Freq}(\mu_j)} & \text{for } 0 < i \leq n \\
        0 & \text{for } i > n \text{ or } i \leq 0
    \end{cases} \]
\end{definition}

In Definition~\ref{def:multi_prob}, the function {\sc Freq}($\mu_j)$ represents the number of observations in which $j$ instances of activity $\mu$ are executed concurrently. As $i$ decreases from $n$ to $1$, the inverse cumulative probability {\sc MDPD}$_n(\mu_i)$ increases, reflecting that if resources can handle $i$ tasks concurrently, they can indeed handle less than $i$, i.e., $i-1, i-2, ..., 1$. Therefore, the probability of handling $i$ concurrent tasks includes all scenarios where the resource also handles $i + 1, ..., n$ tasks. For example, assume activity $\mu$ is executed with the following frequencies: 1 single instance 15 times, two concurrent instances 15 times, three concurrent instances 5 times, and four concurrent instances 5 times. Then, the total number of executions of $\mu$ is 40 ($15 + 15 + 5 + 5$), and let $n = 4$. The cumulative probabilities ({\sc MDPD}$_4$) are the following: for $i = 4$, {\sc MDPD}$_4(\mu_4) = \frac{5}{40} = 0.125$; for $i = 3$, {\sc MDPD}$_4(\mu_3) = \frac{5 + 5}{40} = 0.25$; for $i = 2$, {\sc MDPD}$_4(\mu_2) = \frac{5 + 5 + 15}{40} = 0.625$; and for $i = 1$, {\sc MDPD}$_4(\mu_1) = \frac{5 + 5 + 15 + 15}{40} = 1.0$. For $i > 4$ or $i \leq 0$, the probability is 0.

 Resource Multitasking Capacities (Definition~\ref{def:res_multi_capacities}) rely on the {\sc MDPD} to model how resources can handle multiple activities at the entire process level (globally) or within specific time intervals (locally). In the global multitasking capacity model, the probabilities of a resource performing multiple tasks concurrently are not dependent on the current time granule, whereas in the local multitasking capacity model, these probabilities depend on the current time granule.

\begin{definition}[Resource Multitasking Capacities] \label{def:res_multi_capacities}
Resource Multitasking Capacities are associated with Multitasking Discrete Probability Distributions (MDPD) and fall into the following two categories: 

\begin{itemize}
    \item {\bf Global Multitasking Capacity, $G_{MULT}$:} is a direct one-to-one mapping to an {\sc MDPD}$_n$, modeling the capability of a resource to execute multiple tasks concurrently, i.e., considering the count of activities currently under execution, with each level of activity concurrency uniquely defined within the {\sc MDPD}$_n$.

    \item \textbf{Local Multitasking Capacity,  $L_{MULT}$:} is defined in the context of time granularities. Given a set of recurring slots $\Omega$ and a time granularity $\Delta$, the Local Multitasking Capacity is a function $M: \Omega \times \Delta \to \text{MDPD}_n$, which maps each pair $(\omega \in \Omega, \delta \in \Delta)$ to a Multitasking Discrete Probability Distribution ({\sc MDPD}$_n$). Each pair, referred to as a local multitasking granule, is characterized by the specific {\sc MDPD}$_n$ representing the resource's multitasking capacity within the time granule $\delta$ under the conditions defined by $\omega$. The recurring slots $\omega \in \Omega$ are uniquely identified and bounded by the timestamps $[\tau_0, \tau_{n+1}]$ in $\Delta$.
    
    \item {\sc CanMultiTask} is a function that, given an {\sc MDPD}$_n$ and the current number $\bar{\mu_i}$ of allocated activities to a resource $r$, checks if the resource exceeds their maximum multitasking capacity and, if not, applies a Bernoulli distribution to decide whether the resource can receive (or not) a new one: ({\tt True} {\bf if} $\bar{\mu_i} < \mu_n$) {\bf and} ({\tt True} with {\sc MDPD}$_n$($\bar{\mu_{i+1}})$, {\tt False} with 1 - {\sc MDPD}$_n$($\bar{\mu_{i+1}})$).

\end{itemize}
\end{definition}

To illustrate the concepts described in Definitions~\ref{def:multi_prob} and~\ref{def:res_multi_capacities}, let us focus on a hospital setting where nurses are assigned various patient care tasks, with the workload fluctuating throughout the day. Here, the Global Multitasking Capacity ($G_{MULT}$) gives a general overview of nurses' abilities to handle multiple tasks, like attending to several patients concurrently. This broad assessment does not factor in specific times of the day. For example, Nurse A might be highly capable (as indicated by their {\sc MDPD}$_n$) of managing up to three patients at any time, representing a consistent multitasking ability under typical conditions. 

On the other hand, Local Multitasking Capacity ($L_{MULT}$) introduces the concept of time granularity, e.g., splitting the day into specific intervals or granules, each with varying workload intensities. For example, a nurse may demonstrate a higher multitasking capacity during the busy morning hours, capable of handling more patients. In contrast, another nurse might exhibit greater multitasking efficiency during quieter evening shifts. These differences may depend on individual nurse expertise, peak performance times, and patient flow during specific shifts. Therefore, $L_{MULT}$ dynamically assesses the probability of a nurse managing multiple tasks in different periods, reflecting the fluctuating workloads. 

Table~\ref{tbl:global_local_mult} graphically illustrates Definitions~\ref{def:multi_prob} and~\ref{def:res_multi_capacities}, showcasing the multitasking capabilities of three nurses. In the $G_{MULT}$ section on the left, the varying shades of blue in each row ($\mu_1$, $\mu_2$, $\mu_3$) correspond to the multitasking probabilities for nurses A, B, and C. While Nurses A and C show the capability for up to three concurrent activities, Nurse B demonstrates a high probability of managing one task ($\mu_1$), a moderate likelihood for two ($\mu_2$), but no probability of handling three tasks ($\mu_3$). Each row, thus, graphically encapsulates a single MDPD for an individual nurse, reflecting their overall multitasking capabilities within the Global model.

\begin{table}[tp]
\centering
\begin{minipage}{.49\linewidth}
\centering
\begin{tabularx}{\linewidth}{l @{\hspace{1.0em}} *{3}{>{\centering\arraybackslash}X}}
\toprule
& \multicolumn{3}{c}{$G_{MULT}$} \\
\cmidrule(l){2-4}
 & \multicolumn{3}{l}{} \\
\cmidrule(l){1-4}
\cellcolor{gray!20} & \cellcolor{blue!60}\(\mu_1\) & \cellcolor{blue!40}\(\mu_2\) & \cellcolor{blue!20}\(\mu_3\) \\
\cellcolor{gray!20} {\bf Nurse A} & \cellcolor{blue!60}1.0 & \cellcolor{blue!40}0.7 & \cellcolor{blue!20}0.4 \\
\cellcolor{gray!20} {\bf Nurse B} & \cellcolor{blue!60}1.0 & \cellcolor{blue!40}0.5 &  \\
\cellcolor{gray!20} {\bf Nurse C} & \cellcolor{blue!60}1.0 & \cellcolor{blue!40}0.8 & \cellcolor{blue!20}0.6 \\
\bottomrule
\end{tabularx}
\end{minipage}
\hfill
\begin{minipage}{.49\linewidth}
\centering
\begin{tabularx}{\linewidth}{@{}l @{\hspace{0.01em}} *{9}{>{\centering\arraybackslash}X}@{}}
\toprule
\multicolumn{9}{c}{$L_{MULT}$} \\
\midrule
& \multicolumn{3}{c}{\textbf{Morning}} & \multicolumn{3}{c}{\textbf{Afternoon}} & \multicolumn{3}{c}{\textbf{Night}} \\
\midrule
& \cellcolor{green!60}\(\mu_1\) & \cellcolor{green!40}\(\mu_2\) & \cellcolor{green!20}\(\mu_3\) & \cellcolor{yellow!80}\(\mu_1\) & \cellcolor{yellow!50}\(\mu_2\) & \cellcolor{yellow!20}\(\mu_3\) & \cellcolor{red!60}\(\mu_1\) & \cellcolor{red!40}\(\mu_2\) & \cellcolor{red!20}\(\mu_3\) \\
& \cellcolor{green!80}1.0 & \cellcolor{green!40}0.6 & \cellcolor{green!20}0.3 & \cellcolor{yellow!80}1.0 & \cellcolor{yellow!50}0.5 & \cellcolor{yellow!20}0.2 & \cellcolor{red!60}1.0 & \cellcolor{red!40}0.4 & \cellcolor{red!20}0.1 \\
& \cellcolor{green!60}1.0 & \cellcolor{green!40}0.5 &                         & \cellcolor{yellow!80}1.0 &                          &                          & \cellcolor{red!60}1.0 &  &  \\
& \cellcolor{green!60}1.0 & \cellcolor{green!40}0.7 & \cellcolor{green!20}0.5 & \cellcolor{yellow!80}1.0 & \cellcolor{yellow!50}0.4 & \cellcolor{yellow!20}0.3 & \cellcolor{red!60}1.0 & \cellcolor{red!40}0.5 & \cellcolor{red!20}0.2 \\
\bottomrule
\end{tabularx}
\end{minipage}
\caption{Graphical Representation of Global and Local Multitasking Capacity.}
\label{tbl:global_local_mult}
\end{table}

The $L_{MULT}$ section of Table~\ref{tbl:global_local_mult} provides a detailed view of each nurse's multitasking capacity, differing from the $G_{MULT}$ section by offering multiple MDPDs per nurse – one for each part of the day. Represented by varying colors, each shift or granule – morning (green), afternoon (yellow), and night (red) – has its distinct MDPD. For example, while Nurse A shows multitasking capacity ($\mu_1$, $\mu_2$, $\mu_3$) across the whole day, Nurse B's capacity to handle two tasks ($\mu_2$) appears only in the morning granule of $L_{MULT}$. This distinction reveals that though Nurse B is generally capable of multitasking according to $G_{MULT}$, in practical scenarios, this ability is restricted to mornings, as shown in $L_{MULT}$. The absence of multitasking probabilities for Nurse B in the afternoon and evening in $L_{MULT}$ points to a limitation of the Global model: it does not account for time-specific variations in workload or performance. In contrast, while the Local model provides detailed, time-bound insights, it might not fully capture a nurse's potential to adapt to sudden changes in workload outside fixed time frames. 

Although the global multitasking model could intuitively be seen as a particular case of the local model, this approach is not conceptually accurate. The global model provides a time-agnostic overview of a resource’s capacity to handle multiple tasks concurrently, offering a broad understanding without the need for time-specific data. This abstraction is useful for high-level analysis where time intervals are not a primary concern. Conversely, the local multitasking model incorporates time granularity, capturing variations in multitasking capabilities over different periods. This allows for a more detailed analysis of how resource capacities vary throughout time or across different operational contexts. The distinction between the two models provides flexibility, scalability, and clarity. Practitioners can choose the global model for simplicity or the local model for detailed analysis. The global model suits scenarios without detailed temporal data, while the local model offers time-specific insights. Besides, explicitly differentiating the models helps avoid misinterpretation and ensures the appropriate abstraction is used.

Finally, function {\sc CanMultiTask} in Definition~\ref{def:res_multi_capacities} determines whether a resource can receive a new activity for execution given its current workload. Note that this function does not distinguish between local and global multitasking capacities. Instead, it directly relies on the probabilistic model defined by a MDPD. Therefore, when used over the $L_{MULT}$ model, it requires extracting the corresponding MDPD associated with a granule in $L_{MULT}$.

\subsection{Probabilistic Resource Allocation}

The practical application of the probabilistic models described above relies on the capabilities of a simulation engine to interpret and apply these components to produce the corresponding simulated event logs. Due to their technical nature and space constraints, discussing the simulation engine implementation in detail is beyond the scope of this paper. Still, in the following, we provide some critical insights to consider when developing the simulation engine regarding resource allocation in a probabilistic setting. For further implementation details, we kindly request the readers to check the code repositories whose links are provided later in Section~\ref{sect:evaluation}.

Incorporating probabilistic models into a simulation engine raises challenges due to their stochastic nature. Unlike deterministic functions that produce consistent outcomes, probabilistic models generate varying results in each run. While reflecting the real-world uncertainties, this variability plays against the predictability and repeatability of the simulation outcomes. Additionally, multitasking within these models may result in issues like a single resource monopolizing the execution and skewing simulation results. For example, if the system consistently favors specific resources that complete activities earlier, it can create a self-reinforcing cycle by continuously choosing these resources for new allocations. This may lead to underutilizing other resources and transforming preferred  (overused) resources into bottlenecks. Therefore, the simulation engine requires fine-tuned operation to mitigate imbalances and ensure fair resource allocation.





Aligned with the issues described above, Algorithm~\ref{algo:res_alloc} illustrates the overall semantics for allocating a resource to an enabled event $e$ when simulating a process with resource calendars modeled by Definitions~\ref{def:diff}-~\ref{def:res_availability} and multitasking capacities according to Definitions~\ref{def:multi_prob}-~\ref{def:res_multi_capacities}. In Algorithm~\ref{algo:res_alloc}, all the resources are stored by the following (closest) datetime in which they will be available (or operational) in a priority queue {\sc rQ}. To prevent the overutilization of any single resource in multitasking environments, lines 2 to 7 set up a list of potential resource candidates ({\tt rCandidates}) based on their availability as indicated by the priority queue ({\sc rQ}). So, each resource available before the activity waiting for allocation is eligible, i.e., at each iteration, line 4 retrieves (from the queue) the earliest active resource, $r$, alongside its corresponding available datetime, $\sigma$. 

\begin{algorithm}[tp]
\begin{algorithmic}[1]
\scriptsize
\Function{AllocateResource}{e: {\sc EnabledEvent}, {\sc rQ}: {\sc PriorityQueue}, pt: {\sc ProcessingTime}}

\State rCandidates $\gets$ $\emptyset$
\While{{\bf Exist Resource Candidates}}
    \State r, $\sigma$ $\gets$ {\sc popMin}({\sc rQ})
    \If{$\sigma$ $>$ enabledAt[e] }
        \State {\bf break}
    \EndIf
    \State {\sc Add}(rCandidates, [r, $\sigma$]) 
\EndWhile

\If{rCandidates $\neq$ $\emptyset$}
    \State r, $\sigma$ $\gets$ {\sc RandomUniform}(rCandidates)
    \For{[r', $\sigma'$] $\in$ rCandidates, $r' \neq r$} 
        \State {\sc Enqueue}({\sc rQ}, r', $\sigma'$)
    \EndFor
\Else
    \State r, $\sigma$ $\gets$ {\sc popMin}({\sc rQ})
\EndIf

\If{enabledAt[e] $>$ $\sigma$ }
    \State $\sigma$ $\gets$ {\sc nextAvailableTime}($\sigma$)
\EndIf

\State startedAt[e] $\gets$ $\sigma$
\State completedAt[e] $\gets$ {\sc adjustProcessingTime}($\sigma$, pt)

\If{{\sc CanMultiTask}(r, $MDPD_n^r$, $\bar{\mu}^r$)}
    \State {\sc Enqueue}({\sc rQ}, r, $\sigma$)
\Else
    \State {\sc Enqueue}({\sc rQ}, r, {\sc nextAvailableTime}({\sc Max}$\bar{\mu}^r$completedAt[e]))
\EndIf
\EndFunction

\end{algorithmic}
\caption{Probabilistic Resource Allocation}
\label{algo:res_alloc}
\end{algorithm}

Lines 8-13 of Algorithm~\ref{algo:res_alloc} handle resource allocation. From the list of eligible candidates, a resource is randomly chosen for task assignment, ensuring each candidate has an equal chance (line 9). The unselected resources are then returned to the queue, keeping their current availability datetime (lines 10-11). If no candidate exists, e.g., if no resource was enabled when the activity is ready for allocation, the algorithm assigns the task to the resource at the front of the queue, which is determined by the earliest future availability (lines 12-13). Although $r$ is operational at $\sigma$, the event enablement datetime may occur in a granule posterior to the one of $\sigma$. Therefore, lines 14-15 ensure the resource is not assigned (and starts) a task before its enablement. Then, line 17 computes the event completion datetime by adjusting the preliminary (under ideal conditions) processing time $pt$ as $r$ might be unavailable on some of the intervals spanned by $pt$. 

For efficiency, Algorithm~\ref{algo:res_alloc} does not track the completion of each activity in the execution queue because it adds overhead without it being necessary for task allocation since the scheduling relies on when new activities can start, not when current ones finish. Accordingly, the algorithm checks and updates the execution queue only when a new activity is enabled and then allocated to a resource. Then, at this point, all execution details, including completion times, are computed.

The decision-making process in the execution (priority) queue (lines 18-21) involves determining whether to update the resource's current availability. Accordingly, two possible scenarios can occur when reinserting $r$ into the queue {\sc rQ}. 
In the first scenario, lines 18-19, if the resource still has the capacity to take on a new activity instance, its availability datetime is left unchanged, indicating readiness to accept additional activities immediately. In line 18, the input parameters $MDPD_n^r$ and $\bar{\mu}^r$ of function {\sc CanMultiTask} correspond to the resource capability and the current number of activities under execution by the resource $r$. Although omitted in the pseudocode, the value of $\bar{\mu}^r$  is dynamic and requires to be updated every time a resource is appointed (lines 8-12) and released (lines 18-21). The value of $MDPD_n^r$ remains static across the simulation and can be straightforwardly parsed from the simulation model.


In the second scenario (lines 20-21), the resource cannot take on further activities due to having reached its multitasking capacity or an unsuccessful probabilistic evaluation based on its current workload. One approach is to mark the resource as available as soon as it has some spare capacity again (i.e., as soon as any of the allocated activity instances are completed). This approach would require us to keep track of the completion time of every activity instance in the execution queue and to perform operations on all these completion times. Alternatively, we can set the resource’s next available time in the queue to be equal to the completion time of the last activity instance that was allocated to it. This latter approach does not require us to keep track of the completion time of every activity in the execution queue. Instead, when we detect that allocating an activity to a resource makes it reach its multitasking capacity, we compute the completion time of this activity instance and set the next availability time of the resource accordingly. Since we do not keep track of all activity completion times, we opt for this second approach. Specifically, the algorithm reinserts $r$ into the queue {\sc rQ} with a new datetime corresponding to the next available time after completing all allocated activities. The notation {\sc Max}$\bar{\mu}^r$completedAt[e] indicates the latest completion datetime among all concurrent activities, and $\bar{\mu}^r$ completion times must be dynamically updated.


Note that function {\sc nextAvailableTime} is stochastic, which implies it may return different granules if invoked several times from the same datetime. Therefore, since the resources in {\sc rQ} are already marked as available, it is not advisable to invoke the functions {\sc nextAvailableTime} or {\sc isAvailable} for the same granule because they execute a coin flip again, potentially altering the status of the granule from available to unavailable and vice versa. Similar constraints apply to function {\sc CanMultiTask}, which is also stochastic.

\section{Discovering Probabilistic Resource Calendars}
\label{sect:discovery}

\subsection{Discovering Probabilistic Availability and Performance}

Algorithms \ref{algo:prob_int_discovery}-\ref{algo:update_gr} describe our approach to discovering differentiated probabilistic calendars, which model resource availability from an event log. This approach considers the datetime when tasks are enabled and executed and the resources implicated in each task. To accomplish this, Algorithm~\ref{algo:prob_int_discovery} receives an event log $L$, a time duration $d'$ in minutes, and an angle $\beta \in [0, 1]$ as inputs. The angle $\beta$ will be critical in determining the probabilities within each granule. Initially, lines 2-6 map every resource within $L$ to a granularity $\Delta$, starting at $\delta_0$ = 0 (equating to midnight) with the input duration $d'$ and a total number of granules $n = 1440 // d'$, i.e., spanning the entire day (1440 minutes). For example, if $d'$ is 60 minutes, the probabilistic calendars of each resource will have 24 granules (i.e., $\Delta = \delta_0,..., \delta_{23}$) starting at midnight for every day of the week. More specifically, the matrices $\lambda$ and $\Lambda$ tally, respectively, the frequency at which a resource $r$ was detected to be operational ($\lambda$) and when it was required ($\Lambda$) within a specific time granule $\delta$ on a day of the week $\omega$ in the log $L$. The matrix $M$ counts the frequency at which the most utilized resource was operational within each p-granule.

\begin{algorithm}[tp]
\begin{algorithmic}[1]
\scriptsize
\Function{DiscoverIntervals}{$L$: {\sc EventLog}, d': Minutes, $\beta$: angle}

\For{{\bf each} day of week $\omega \in \{Monday, ..., Sunday\}$}
    \State $M$[$\omega$] $\gets$ $\Delta(\tau_0 = 0, d = d', n = 1440 // d')$
    \For{{\bf each} resource $r \in L$}
        \State $\lambda$[$r$][$\omega$] $\gets$ $\Delta(\tau_0 = 0, d = d', n = 1440 // d')$
        \State $\Lambda$[$r$][$\omega$] $\gets$ $\Delta(\tau_0 = 0, d = d', n = 1440 // d')$
    \EndFor
\EndFor

\For{ {\bf each} trace $T$ $\in$ $L$}
\State {\sc ComputeEnablingTimes}($T$)
    \For{ {\bf each} event $e$ $\in$ $T$}
        \State {\sc Trapezoidal}(e, $\lambda$, $\Lambda$, $M$, $\beta$, False)
        \State {\sc Trapezoidal}(e, $\lambda$, $\Lambda$, $M$, $\beta$, True)
    \EndFor
\EndFor

\For{{\bf each} resource $r \in L$}
    \For{{\bf each} day of week $\omega \in \{Monday, ..., Sunday\}$}
        \For{{\bf each} time granule $\delta$ $\in$ $\Delta(\tau_0 = 0, d = d', n = 1440 // d')$}
            \State $P_{ABS}$ $\gets$ $\lambda$[$r$][$\omega$][$\delta$] / $\Lambda$[$r$][$\omega$][$\delta$]
            \State $P_{REL}$ $\gets$ $\lambda$[$r$][$\omega$][$\delta$] / $\Lambda$[$\omega$][$\delta$]
        \EndFor
    \EndFor
\EndFor
\State {\bf return} $P_{ABS}$, $P_{REL}$
\EndFunction

\end{algorithmic}
\caption{Discovery of Probabilistic Calendars}
\label{algo:prob_int_discovery}
\end{algorithm}

\begin{algorithm}[tp]
\begin{algorithmic}[1]
\scriptsize
\Function{Trapezoidal}{$e$, $\lambda$, $\Lambda$, $M$, $\beta$, allocated}

\If{allocated}
    \State $GR$ $\gets$ {\sc ExtractTimeGranules}(startedAt[$e$], completedAt[$e$])
\Else
    \State $GR$ $\gets$ {\sc ExtractTimeGranules}(enabledAt[$e$], startedAt[$e$])
\EndIf

\If{$|GR| = 1$}
    \State {\sc UpdateGranules}(1.0, $e$, $\delta_0 \in GR$, $\delta_0 \in GR$, $\lambda$, $\Lambda$, $M$, allocated)
\Else
\State $p \gets 1.0$
\State $f \gets$ 1.0 / ($|GR|$ // 2) * $\beta$) {\bf if} $\beta$ $>$ 0 {\bf else} 1.0

\State $s \gets 0$, $e \gets$ $|GR| - 1$
\While{$\delta_s < \delta_e$}
    \State {\sc UpdateGranules}($p$, $e$, $\delta_s \in GR$, $\delta_e \in GR$, $\lambda$, $\Lambda$, $M$, allocated)
    \State $p \gets p - f$
    \State $s \gets s + 1$, $e \gets e - 1$
\EndWhile
\EndIf
\EndFunction

\end{algorithmic}
\caption{Availability Calculation- Trapezoidal Method }
\label{algo:trapezoidal}
\end{algorithm}

Subsequently, Algorithm~\ref{algo:prob_int_discovery} in lines 7-11 computes the enabling times for each event within each trace $T$ in the log and applies the trapezoidal method as detailed in Algorithm~\ref{algo:trapezoidal} to every event $e$. Finally, in lines 12-16, for each resource in the event log and each day of the week, Algorithm~\ref{algo:prob_int_discovery} calculates the absolute and relative probabilities for each time granule, which are returned later in line 17. Specifically, the absolute probability $P_{ABS}$ computes the ratio of the frequency at which a resource was operational to the total frequency at which it was required for an activity. Note that the required intervals also count the instances when the resource was operational. The relative probability $P_{REL}$ measures the ratio between $\lambda$ but divided by the most frequent resource in the corresponding granule. As such, the most frequently engaged resource in a given p-granule would get a relative probability of 1.

Algorithm~\ref{algo:trapezoidal} describes a trapezoidal method to measure the resource availability within p-granules related to a given event, distributing the availability in a trapezoidal shape over the event duration. So, p-granules containing the event's start and end datetimes check the resource as operational. In contrast, the resource status is vague in the remaining granules between these points. The trapezoidal method tackles this uncertainty by assigning a weight of 1.0 to granules matching an event's start and end datetimes. Then, it reduces the weight of resource availability in proportion to the time distance of a p-granule from the empirically confirmed operational p-granules.

Algorithm~\ref{algo:trapezoidal} takes six parameters as input: an event $e$, the matrices $\lambda$, $\Lambda$ and $M$, to be updated, the angle $\beta$ shaping the trapezoid, and a boolean variable, {\it allocated}, which defines two types of intervals. The first type (False) corresponds to the interval between enabling and starting times of the event, a period in which the resource is required but not yet available. The second type (True) refers to the interval between starting and ending times of the event, a period in which the resource might be operational executing the related task. 

Algorithm~\ref{algo:trapezoidal} operates as follows: lines 2-5 extract the sequence of granules according to the {\it allocated} parameter types. Then, lines 6-15 differentiate two cases to update the availability. If the duration of the event spans only one granule, then it is weighted twice with 1.0, as the resource started and ended the event within it (lines 6-7). When the event spans multiple granules, the initial weight is 1.0 at the boundaries and decreases towards the center (lines 9-15). The rate of this decrement relies on the angle $\beta$ as described by the formula on line 10. A $\beta = 0.0$ assigns a weight of 1.0 to the boundary granules and 0.0 to the remaining ones. In contrast, a $\beta = 1.0$ decreases the weight of each granule in the sequence by a factor $f$ given by $1.0 / (|GR| // 2) * \beta$). For example, given five granules, $\beta = 0.0$ results in [1.0, 0.0, 0.0, 0.0, 1.0], and $\beta = 1.0$ in [1.0, 0.5, 0.0, 0.5, 1.0].

\begin{algorithm}[tp]
\begin{algorithmic}[1]
\scriptsize
\Function{UpdateGranules}{$p$, $e$, $\delta_s$, $\delta_e$, $\lambda$, $\Lambda$, $M$, allocated}
    \For{$\delta \in$ $\{\delta_s, \delta_e\}$}
        \State $\omega$ $\gets$ {\sc DayOfWeek}($\delta$, $e$)
        \For{{each} $rCand \in$ {\sc taskResources}[task[$e$]]}
            \If{{\bf not} {\sc isBusy}[$rCand$][{\sc Datetime}($\delta$)][$\delta$]}
                \State $\Lambda$[$rCand$][$\omega$][$\delta$] $\gets +1$
            \EndIf
        \EndFor
        \If{allocated}
            \State $\lambda$[resource[$e$]][$\omega$][$\delta$] $\gets +p$
            \State $\Lambda$[resource[$e$]][$\omega$][$\delta$] $\gets +1$
            \State $M$[$\omega$][$\delta$] $\gets$ {\sc max}($M$[$\omega$][$\delta$], $\lambda$[resource[$e$]][$\omega$][$\delta$])
        \EndIf
    \EndFor
\EndFunction

\end{algorithmic}
\caption{Update Time Granule Availability Weights}
\label{algo:update_gr}
\end{algorithm}

Algorithm~\ref{algo:update_gr}, called in lines 7 and 13 of Algorithm~\ref{algo:trapezoidal}, updates the granules determined by the trapezoidal method. This algorithm requires eight input parameters: the availability weight $p$, computed from $\beta$, two granules $\delta_s$ and $\delta_e$ that are at equal distance from the interval boundaries, the matrices $\lambda$, $\Lambda$, and $M$, and the boolean variable {\it allocated}.

For each of the two granules $\delta$ received, Algorithm~\ref{algo:update_gr} proceeds as follows: Line 3 identifies the day of the week on which $\delta$ occurs. Subsequently, the loop in lines 4-6 increments by one the frequency of pair $<\omega, \delta>$ in the matrices $\Lambda$ for each non-operational resource capable of performing the task defined by the event $e$. This adjustment relies on the idea that all available resources potentially suited to the task were considered eligible but were ultimately not selected. This modification applies to granules within both interval types, from the event's enablement to start and from start to end. Then, lines 7-10 update p-granules within the intervals where a resource might be operational, i.e., engaged in the event execution from start to end. Thus, the matrix $\lambda$ of the resource executing $e$ is updated with the estimated availability factor $p$. The frequency of $<\omega, \delta>$ in the matrix $\Lambda$ increments by 1, acknowledging that the resource was indeed required. Finally, the algorithm checks and updates whether a new, most frequent resource has emerged for that p-granule in matrix $M$.

When processing the event log, the allocated resources would ideally work on the corresponding tasks without interruption. However, according to their calendars, they can have some resting time in between. Accordingly, Algorithm~\ref{algo:fit_processing} recalibrates the processing times observed in the event log to match granular time intervals identified by the trapezoidal method. The input of the algorithm consists of the event log $L$, the absolute and relative probability functions built by Algorithms~\ref{algo:prob_int_discovery}-\ref{algo:update_gr}, and a float number $\kappa$ referring to the minimum frequency threshold for a resource-task pair occurrence needed to estimate their adjusted processing times. 

This paper follows the differentiated performance model presented in~\cite{Lopez-PintadoD22}. In this model, each individual resource has a different performance. Thus, the processing times are calculated for each pair (resource, task) in $L$. Accordingly, lines 2-3 of Algorithm~\ref{algo:fit_processing} initialize the $adjTimes$ matrix to tally the adjusted times and group the events in $L$ by resource-task pairs. Lines 4-14 iterate over each event $e$ for each pair resource $r$, task $t$. First, it extracts the granules $\delta_i$ from the start to the end of $e$ (i.e., representing the processing time) and identifies each granule's corresponding weekday $\omega$. Subsequently, the adjusted processing time is the sum of each granule duration ($d_{\delta_i}$), each multiplied by their maximum probability between $P_{ABS}$, $P_{REL}$. The rationale is that granules are allocated selectively from their probabilities. Thus, multiplying each granule duration by the corresponding probability factor maintains or decreases the overall processing time, which, on average, may converge to the actual operational times. 
The first and last granules might only be partially covered. Thus, the notation $(\delta_i, e)_i$, with $i = 0, n$ represents the actual durations from the event's start to the end of the first granule and from the start of the last granule to the event's end.

\begin{algorithm}[tp]
\begin{algorithmic}[1]
\scriptsize
\Function{FitProcessingTimes}{$L$: EventLog, $P_{ABS}$, $P_{REL}$, $\kappa$}
\State $adjTimes$ $\gets$ $[resources \in L, tasks \in L]$
\State $resourceTaskEvents$ $\gets$ {\sc GroupEventsByPairResourceTask}($L$)
\For{{\bf each} $r, t \in resourceTaskEvents$}
    \For{{\bf each} event $e \in resourceTaskEvents[r][t]$}
        \State $GR$ $\gets$ {\sc ExtractTimeGranules}(startedAt[$e$], completedAt[$e$])
        \State $pTime$ $\gets$ $0$
        \For{{\bf each} $\delta_i \in GR$}
            \State $\omega$ $\gets$ {\sc DayOfWeek}($\delta$, $e$)
            \If{$i = 0$ {\bf or} $i = n$}
                \State $d'$ $\gets$  +$(\delta_i, e)_i * \max(P_{ABS}[r][\omega][\delta_i], P_{REL}[r][\omega][\delta_i])$
            \Else
                \State $d'$ $\gets$ +$d_{\delta_i} * \max(P_{ABS}[r][\omega][\delta_i], P_{REL}[r][\omega][\delta_i])$
            \EndIf
            \State {\sc Append}($adjTimes[r][t]$, $d'$)
        \EndFor
        
    \EndFor
\EndFor
\State $pTimeDistr$ $\gets$ $[resources \in L, tasks \in L]$
\For{{\bf each} $r, t \in resourceTaskEvents$}
\If{$|adjTimes[r][t]| \geq \kappa$}
    \State $pTimeDistr[r][t]$ $\gets$ {\sc BestFitDistribution}($adjTimes[r][t]$)
\Else
    \State $pTimeDistr[r][t]$ $\gets$ $\emptyset$
\EndIf
\EndFor

\For{{\bf each} $r, t \in resourceTaskEvents$}
\If{$pTimeDistr[r][t] = \emptyset$}
    \State $rCandidate$ $\gets$ {\sc FindClosestMeanCandidate}($t$)
    \If{$rCandidate \neq \emptyset$}
     \State $pTimeDistr[r][t]$ $\gets$ $pTimeDistr[rCandidate][t]$
     \Else 
        \State  $pTimeDistr[r][t]$ $\gets$ {\sc BestFitDistribution}($\forall$ $t \in L$)
    \EndIf
\EndIf
\EndFor
\State {\bf return} $pTimeDistr$
\EndFunction
\end{algorithmic}
\caption{Fit Processing Times to Probabilistic Granules}
\label{algo:fit_processing}
\end{algorithm}

After adjusting the processing times observed in $L$ according to the probabilistic calendars, Algorithm~\ref{algo:fit_processing} calculates a distribution function to model them. In this regard, for every resource-activity pair (lines 15-20), the algorithm ensures the number of processing times calculated fulfills the minimum level of significance settled by the parameter $\kappa$. If this condition is satisfied, the function {\sc BestFitDistribution} constructs a histogram from them. It employs curve-fitting techniques to identify a probability distribution that offers the most precise approximation to the histogram, i.e., the one with the lowest residual sum. In cases where the pairs do not meet the $\kappa$ requirement, the distribution function is built by aggregation (lines 21-27). Specifically, if other resources with an associated distribution executing the corresponding task exist, the algorithm assigns the distribution function of that resource with the closest mean to the non-adjusted processing times. Otherwise, the distribution aggregates the processing times observed for all the resources that performed the task.

\subsection{Discovering Probabilistic Multitasking Capacity}

Algorithms \ref{algo:global_capacity} and ~\ref{algo:local_capacity} describe our approach to discovering, from an event log, the probabilistic global and global multitasking capacity previously formalized in Definition~\ref{def:res_multi_capacities}. Multitasking capacity is a concept directly linked to resources as it models their ability to handle multiple activities simultaneously. Accordingly, both Algorithms~\ref{algo:global_capacity} and ~\ref{algo:local_capacity}, initially in line 2, groups the events in the log $L$ received as input by the resources executing them, i.e., retrieving a mapping of each resource to their list of the executed events. 

For each resource, $r$, loop in lines 4-15, Algorithm~\ref{algo:global_capacity} splits the events by their start and end states (line 5), which is required to track activities' beginning and completion and identify overlapping activities. These events are further sorted chronologically (line 6), setting the ground for applying a sweep-line strategy (lines 9-14). By processing events chronologically, the algorithm can `sweep' through time, making it possible to analyze the overlap of activities in a temporal context.

\begin{algorithm}[tp]
\caption{Calculate Global Multitask Capacity}
\label{algo:global_capacity}
\begin{algorithmic}[1]
\scriptsize
\Function{Discover$G_{MULT}$}{$L$: Log}

\State $resourceEvts$ $\gets$ {\sc GroupEventsByResources}($L$)
\State $G_{MULT}$ $\gets$ $\emptyset$

\For{{\bf each} resource $r$ $\in$ $resourceEvts$}
    \State $rEvts$ $\gets$ {\sc SplitEventsByStartEndStates}($resourceEvts$[$r$])
    \State $sortedEvts$ $\gets$ {\sc SortEventsByDatetime}($rEvts$)
    \State $activeEvts$ $\gets$ $\emptyset$ 
    \State $M$ $\gets$ $\{\}$ \Comment{\textcolor{blue}{Dictionary $<\mu_i: F_\mu>$}}
    \For{{\bf each} event $e$ $\in$ $sortedEvts$}
        \If{{\sc State}[$e$] $=$ `$Start$'}
            \State {\sc Add}($activeEvts$, $e$)
            \State $M$[$\mu_i$] $\gets$ $M$[$\mu_i$] $+$ $1$, {\bf with} $i$ $=$ $|activeEvts|$
        \Else
            \State {\sc Remove}($activeEvts$, $e$)
        \EndIf
    \EndFor
    \State $\{\mu_1, ..., \mu_n \}$, $F_\mu$ $\gets$ {\sc Keys}($M$), {\sc Values}($M$)
    \State $G_{MULT}$[$r$] $\gets$ {\sc GetMDPD}($\{\mu_1, ..., \mu_n \}$, $F_\mu$, $|sortedEvts| / 2$)
\EndFor

\State {\bf return} $G_{Mult}$  
    
\EndFunction
\end{algorithmic}
\end{algorithm}

Lines 9-14 focus on constructing a multitasking frequency dictionary $M$, represented as $<\mu_i: F_\mu>$. This dictionary tracks multitasking levels ($\mu_i$) and their frequencies ($F_\mu$) for each resource. For example, assume the resource executed groups of 3 tasks concurrently 20 times across the process execution. Here, $M$ will store the pair $<\mu_3, 20>$, i.e., $F_\mu$[3] = 20. As Algorithm~\ref{algo:global_capacity} handles each `start' event within the sorted list (lines 10-12), it updates $M$ by increasing the count at the current level of active tasks. Each `start' event increases the frequency of the respective multitasking level, reflecting the concurrent handling of activities. Conversely, `end' events remove tasks from the active set without altering $M$ (lines 13-14). 

Finally, lines 15-16 in Algorithm~\ref{algo:global_capacity} calculate and map each resource's corresponding Multitasking Discrete Probability Distribution (MDPD) - Definition~\ref{def:multi_prob}, using function {\sc GetMDPD} (presented in Algorithm~\ref{algo:calculate_mdpd} discussed later in this section). This function requires the multitasking levels and frequencies from $M$. Also, it receives half the total number of events reflecting the overall activity volume, i.e., since each event was recorded twice (once at the start and once at the end).

Moving now to the local multitasking calculation, Algorithm~\ref{algo:local_capacity} receives the desired granularity and recurring slots in addition to the event log. Note that this granularity may differ from the one used to model the resource availability calendars. For example, the calendar may split each weekday into 24 hours, while the multitasking capacity could be bound to 3 shifts of 8 hours each day (morning, afternoon, evening), both repeating periodically in recurring slots (cycles) of 7 days, i.e., Monday, Tuesday, ..., Sunday.

\begin{algorithm} [tp]
\caption{Calculate Local Multitask Capacity}
\label{algo:local_capacity}
\begin{algorithmic}[1]
\scriptsize
\Function{Discover$L_{MULT}$}{$L$: Log, $\Delta$: Granularity, $\Omega$: RecurringSlots}

\State $resourceEvts$ $\gets$ {\sc GroupEventsByResources}($L$)
\State $L_{MULT}$ $\gets$ $\emptyset$

\For{{\bf each} resource $r$ $\in$ $resourceEvts$}

    \State $G^D$ $\gets$ $[]$
     \For{{\bf each} event $e$ $\in$ $resourceEvts$[$r$]}
        \State {\sc Add}($G^D$, {\sc ToDate}$\Bar{\Delta\Omega}$($\Delta$, $\Omega$, startedAt[$e$], completedAt[$e$]))
     \EndFor

    \For{{\bf each} pair $<\omega, \delta>$ $\in$ $G^D$}
        \State $sortedG^D$ $\gets$ {\sc SortGranulesByStartDatetime}($G^D$[$\omega$][$\delta$])   
        \State $\{\mu_1, ..., \mu_n \}$, $F_\mu$ $\gets$ {\sc IntersectIntervals}($sortedG^D$[$\omega$][$\delta$])
        \State $L_{MULT}$[$r$][$\omega$][$\delta$] $\gets$ {\sc GetMDPD}($\{\mu_1, ..., \mu_n \}$, $F_\mu$, $|$$G^D$[$\omega$][$\delta$]$|$)
    \EndFor
\EndFor

\State {\bf return} $L_{MULT}$  
    
\EndFunction
\end{algorithmic}
\end{algorithm}

For each resource, $r$, loop in lines 4-15, Algorithm~\ref{algo:local_capacity} iterates through events executed by them (lines 6-7), converting their start and end times into granules that match the input granularity and recurring slots. The d-granules produced by the function {\sc ToDate}$\Bar{\Delta\Omega}$ and stored in the list $G^D$ are similar to the p-granules defined in ~\ref{def:prob_calendar} related to resource calendars. However, unlike p-granules (containing a probability), d-granules capture start/end datetimes. These datetimes specify the exact time intervals of activity occurrences during execution, thereby offering a precise temporal mapping of the extracted granules. For example, assume an event occurs from 27/03/2024 10:15:00 to 27/03/2024 11:45:00. Assuming a granularity splitting the day in granules of 1 hour starting at midnight, it produces the d-granules: [10:00:00,  11:00:00) $\rightarrow$ [27/03/2024, 27/03/2024) and [11:00:00,  12:00:00] $\rightarrow$ [27/03/2024, 27/03/2024). Initially, granularities, as in Definitions~\ref{def:granularity}-~\ref{def:prob_calendar}, do not contain the date component.

The loop in lines 8-11 of Algorithm~\ref{algo:local_capacity} calculates the MDPD following a sweep line strategy, not to count event occurrences as in the global method but to identify overlaps in the granule intervals extracted from the event durations. For each d-granule, the algorithm sorts the intervals by their start times. Then, it calculates their intersections, indicating events that occur concurrently within these granules, thus the existence of multitasking at those specific time intervals, described by the corresponding multitasking levels ${\mu_1, ..., \mu_n}$ and their frequencies $F_\mu$. Finally, the algorithm uses the same {\sc GetMDPD} function as the global approach to calculate MDPD but scoped to each granule. 

\begin{algorithm}[tp]
\caption{Calculate MDPD Model}
\label{algo:calculate_mdpd}
\begin{algorithmic}[1]
\scriptsize
\Function{GetMDPD}{$\{\mu_1, ..., \mu_n \}$, $F_\mu$, $TotalObs$}
\State $MDPD$ $\gets$ $\emptyset$
\State $cumulFreq$ $\gets$ 0
\State $i = n$
\While{$i \geq$ 1}
    \State $cumulFreq$ $\gets$ $cumulFreq$ + $i$ * $F_\mu[i]$
    \State $MDPD$[$\mu_i$] $\gets$ $cumulFreq$ / $TotalObs$
    \State i $\gets$ $i - 1$
\EndWhile
\State {\bf return} $MDPD$
\EndFunction
\end{algorithmic}
\end{algorithm}

Algorithm~\ref{algo:calculate_mdpd} outlines the function for calculating the MDPD invoked from Algorithms~\ref{algo:global_capacity} and~\ref{algo:local_capacity}, using multitasking levels ${\mu_1, ..., \mu_n }$, their frequencies $F_\mu$, and the total number of observations $TotalObs$. The variable $cumulFreq$ captures the principle that a resource capable of handling $n$ concurrent events can manage fewer events. Consequently, the probability of handling fewer concurrent events increases. From the highest multitasking level, the algorithm cumulatively adds the frequency of each level ($i$ times $F_\mu[i]$) to $cumulFreq$, thus aggregating frequencies up to and including that level. As it moves to lower levels, $cumulFreq$ grows, reflecting the increased likelihood of fewer concurrent tasks. Each level's probability is calculated by dividing $cumulFreq$ by $TotalObs$, ensuring higher probabilities correspond to the lower multitasking levels in the MDPD.

\section{Evaluation}
\label{sect:evaluation}

We implemented the proposed approach by extending the {\sc Simod} tool for automated discovery of simulation models from event logs~\cite{CamargoDG20}\footnote{\url{https://github.com/AutomatedProcessImprovement/Simod}} and the {\sc Prosimos} simulation engine~\cite{Lopez-PintadoD22,Lopez-PintadoHD22}\footnote{\url{https://github.com/AutomatedProcessImprovement/Prosimos}} to support probabilistic availability calendars and multitasking capacities. 
Using the extended versions of {\sc Simod} and {\sc Prosimos}, we conducted experiments to address the following questions: 
\begin{itemize}
\item[{\bf I.}] Does the use of probabilistic calendars improve the accuracy of BP simulation models discovered from event logs relative to crisp calendars? 
\item[{\bf II.}] How do the two approaches for discovering BP simulation models with multitasking (global versus local) perform relative to each other and relative to single-task simulation models?
\item[{\bf III}] How does the presence of extended periods of resource unavailability in the input log, not captured in the discovered periodic pattern, affect the accuracy of the discovered models?
\end{itemize}


\subsection{Datasets}

\subsubsection{Datasets for Evaluating the Availability Calendar Models}


The experiments corresponding to Evaluation Question {\bf I.} rely on eight synthetic and three real-life logs. We used a business process simulation model (herein called the {\it Loan} model) as a starting point to generate the synthetic logs.\footnote{This process model and the generated event logs are provided in the reproducibility package referenced at the end of this article.}
This BP simulation model includes the following simulation parameters: branching probabilities for each conditional flow in the process model; case arrival information; distribution of processing times of each activity; a mapping between activities in the process model and resource profiles; and a mapping from each resource profile to a set of resources (every resource profile maps to multiple resources). The case arrival is such that a simulation would generate 2000 cases, which are created at random intervals that follow a Poisson distribution with a mean arrival rate of 20 cases per day of 8 working hours (8:00-12:00 and 13:00-17:00), Monday to Friday, i.e., one new case is generated every 24 minutes between these times of the weekday. Each task-resource pair has a different distribution of processing times chosen randomly.

We produced four variants of the {\it Loan} model by altering the availability calendars of the resources. These four variants are herein designated with the labels ``24'', ``8'', ``8/4'' and ``24/A'', The resource calendars in each of these variants were derived as follows:
\begin{itemize}
    \item 24 -- all resources are available 24 hours/7 days per week.
    \item 8 -- all resources are available on an 8-hour shift from Monday to Friday, with working hours from 8:00-12:00 and 13:00-17:00.
    \item 8/4 -- half the resources work the 8-hour shift (as above) while the other half work part-time (13:00-17:00 Monday to Friday). The resources are randomly divided into these two halves. This scenario is intended to cover scenarios where resources have different availability calendars, potentially impacting the waiting times.
    \item 24/A -- one-third of the resources work the 8-hour shift, one-third work the 4-hour shift, and the remaining third work 24 consecutive hours followed by a 48-hour rest period, and so on. This scenario is a more extreme version of 8/4 scenario. It is intended to validate how the approaches perform when there are major differences in availability calendars. Also, the calendar where resources work 24 hours followed by a 48-hour rest period has a non-circadian periodicity, e.g., resources work on Mondays on some weeks but rest on Mondays on other weeks.
\end{itemize}

Next, for each of the above four model variants, we generated two sub-variants, herein designated with the labels ``B'' (stands for ``Balanced'') and ``U'' (stands for ``Unbalanced''). The balanced variants capture scenarios where all the resources assigned to a given activity A exhibit similar workloads, whereas the unbalanced variants capture scenarios where some resources assigned to a given activity A have a higher workload (i.e., get assigned more activity instances) than other resources who are also assigned to the same activity. The idea here is that in the unbalanced variant, some resources more eagerly take on new activity instances than in others, i.e., some resources are ``eager'' and others are ``lazy''. The eager resources correspond to resources who take on new activity instances all the time (within their availability calendar), whereas the lazy resources sometimes do not start a new activity instance even if they are on duty according to their availability calendar. Thus, the eager resources are likely to exhibit higher probabilities in their probabilistic availability calendars than the lazy ones. We expect that the probabilistic availability calendar approach will perform better in unbalanced scenarios because this approach is able to differentiate between eager and lazy resources, whereas the crisp calendars would not.

We generated one event log from each of the 8 resulting variants of the original {\it Loan} model.
To generate an event log with an unbalanced workload, we needed a simulator that assigns more cases to some resources relative to others. We implemented a trace generator specifically for this purpose. This 
generator takes a BP simulation model as input. 
It first generates the start times of each case according to the case arrival information in the input model. 
Next, the trace generator creates a sequence of activity instances for each case by animating the BPMN model step-by-step (i.e., playing the token game). Along the way, the trace generator uses the branching probabilities associated to the conditional flows of the BPMN model to determine which conditional flow should be taken at each traversal of a decision gateway. When an activity instance is enabled, the trace generator assigns it to a resource. The generator supports two options for the assignment of resources to activity instances: a balanced option and an imbalanced one. In the balanced option, each enabled instance of a given activity A is assigned to one of the resources that can perform activity A in a round-robin manner (so that each resource is assigned a similar number of instances of this activity). In the unbalanced option, the assignment is also done in a round-robin manner but in such a way that resources receive different workloads. Specifically, we divided the resources of each resource profile into 6 groups. Out of every 21 instances that could be allocated to this resource profile, the resources in group 1 received 1 activity instance, those in group 2 received the next 2 activity instances, those in group 3 the next 3 activity instances, and so on.

Once an activity instance in a trace is assigned to a resource, the trace generator calculates its start time by taking into account the time point when the activity instance was enabled (which is recorded during the token game animation), the availability calendar of the assigned resource, and the end times of other activity instances previously assigned to this resource. Given the start time, the end-time of the activity instance is trivially obtained by adding the processing time of the activity instance (drawn randomly from the distribution of processing times associated to the activity).


As a result of the above procedure, we obtained a total of 8 synthetic event logs: 4 balanced and 4 unbalanced. The resulting logs are herein named B-24, B-8, B-8/4, B-24/A (four balanced-workload logs with different availability calendars) and U-24, U-8, U-8/4, U-24/A (for unbalanced-workload logs)

We complement the above 8 synthetic logs with three real-life logs with different characteristics, as per Table~\ref{tbl:log-description}.\footnote{To be able to discover simulation models, every one of the logs in this evaluation needs to record the start time, end time, and resource of each activity instance. This constraint precluded us from using publicly available event logs that do not contain start times.} The first one, namely {\it BPIC-2012}, is an event log of a loan application process at a Dutch financial institution, which was made available as part of the Business Process Intelligence Challenge 2012.\footnote{\url{https://doi.org/10.4121/uuid:3926db30-f712-4394-aebc-75976070e91f}} The second log, namely {\it BPIC-2017} log, is the event log of the 2017 edition of the same challenge.\footnote{\url{https://doi.org/10.4121/uuid:5f3067df-f10b-45da-b98b-86ae4c7a310b}} This second log comes from the same business process as the {\it BPI-2012} log, but it was extracted 5 years later. We preprocessed the BPI-2017 log by following the recommendations of the winning teams of the {\it BPIC-2017} challenge to eliminate outliers and incomplete cases.\footnote{\url{https://www.win.tue.nl/bpi/2017/challenge.html}} The third log, {\it CALL}, is from a call center process. This latter event log includes a high volume of cases with short duration (on average, two activities per case). All the logs are included in the reproducibility package referenced at the end of this paper.

For the real-life logs, we did not have a process model as a starting point (only the log). Given that our approach is independent of the algorithm used for process model discovery, we did not automatically generate the models from these real-life logs. Instead, we discovered the BPMN process models from the event logs using the Apromore process mining tool.


Table~\ref{tbl:log-description} gives descriptive statistics of the logs, including the number of traces, events, activities, and resources. The first two columns list the characteristics of the synthetic logs (the four balanced, Loan-B) and the four unbalanced ones (Loan-U). The remaining columns describe the real-life logs.
   \begin{table}[tp]
    \centering
    \small
\begin{adjustbox}{width=0.99\textwidth}
    \begin{tabular}{{lccccc}}
        \toprule
         & \textbf{Loan-B/} & \textbf{Loan-U/} & \textbf{BPIC12/} & \textbf{BPIC17/} & \textbf{CALL/}  \\
        \midrule
        \textbf{Traces}           & 2000  & 2000  & 8616  & 30276   & 521779 \\
        \midrule
        \textbf{Events}           & 23332 & 23288 & 59301 & 240854  & 900374 \\
        \midrule
        \textbf{Activities}       & 17    & 17    & 6     & 8       & 19     \\
        \midrule
        \textbf{Resources}        & 54    & 54    & 58    & 148     & 3021   \\
        \bottomrule
    \end{tabular}
\end{adjustbox}
   \caption{General characteristics of event logs for availability calendar evaluation.}
   \label{tbl:log-description}
 \end{table}

\subsubsection{Datasets for Evaluating the Multitasking Capacity Models}

The experiments corresponding to Evaluation Question {\bf II} rely on eight synthetic and four real-life logs. The eight synthetic logs were generated starting from the same loan application process model that we used to generate the synthetic logs for the availability calendar evaluation. We only modified the following two steps in the log generation approach to ensure the presence of multitasking in the resulting logs. First, we increased the inter-arrival rate so that each synthetic log contains 2000 cases with inter-arrival times following a Poisson distribution with a mean of 108 cases per day (i.e., twice the number of resources in the process) from Monday to Friday. Additionally, the cases were created in two batches, one at 8:00 and the other at 13:00, instead of being spread across all working hours as we did in datasets to assess the availability calendar. In this way, we ensure that many cases start at once, which increases the changes of the activity instances in these cases overlapping each other.
Second, the start time of each activity instance was altered so that each activity instance is allocated to a resource and started as soon as it becomes enabled (from a control-flow perspective). Thus, any pair of activity instances that are enabled more or less at the same time will overlap, and this increases the chances that multiple activities allocated to the same resource will overlap (multitasking). The remaining parameters related to control flow, allocation of activities to resources, balanced and unbalanced workloads, and processing times were kept as described for the availability calendars evaluation datasets.

\begin{table}[tp]
    \centering
    \small
\begin{adjustbox}{width=0.99\textwidth}
    \begin{tabular}{{lcccccc}}
        \toprule
         & \textbf{Loan-B'/} & \textbf{Loan-U'/} & {\bf AC-CRE} & {\bf GOV} & {\bf WK-ORD} & {\bf PAY}  \\
        \midrule
        \textbf{Traces}        & 2000  & 2000  & 954      & 4856     &  14945    & 31254 \\
        \midrule
        \textbf{Events}        & 23632 & 23472 & 6829     &  22529  & 112821    &  182859 \\
        \midrule
        \textbf{Activities}    & 17    & 17    & 18        &  85    & 24         &  8    \\
        \midrule
        \textbf{Resources}     & 54    & 54    & 561  &  16 &  193  & 9   \\
        \bottomrule
    \end{tabular}
\end{adjustbox}
   \caption{General characteristics of event logs for multitasking capacity evaluation.}
   \label{tbl:log-description-multi}
 \end{table}

 Table~\ref{tbl:log-description-multi} outlines the characteristics of the synthetic logs (the balanced Loan-B' and the unbalanced Loan-U') and those of the real-life logs discussed below. Additionally, Tables~\ref{tbl:multi-synt-log-desc} and~\ref{tbl:multi-real-log-desc} display the multitasking statistics of each event log. These two tables provide the mean and median number of activity instances that a given resource performs concurrently. For example, a mean value of two in one of these tables implies that, on average, when a resource is busy, it performs two activity instances concurrently. The last row in these two tables (the ``Ratio'') displays the percentage of activity instances performed multitasked (i.e., performed by a resource concurrently with at least one other activity instance). 

 \begin{table}[tp]
    \centering
    \small
\begin{adjustbox}{width=0.99\textwidth}
    \begin{tabular}{{lcccccccc}}
        \toprule
                                   & {\bf B'-24/} & {\bf U'-24/} & {\bf B'-8/} & {\bf U'-8/} & {\bf B'-8/4/} & {\bf U'-8/4/} & {\bf B'-24/A/} & {\bf U'-24/A/} \\
        \midrule
        \textbf{Mean}       & 2.61        & 2.9         & 4.69       & 5.32       & 6.39         & 7.4          & 9.13          & 10.28   \\
         \midrule
        \textbf{Median}    & 2.0         & 2.0         & 3.0        & 3.0        & 4.0          & 4.0          & 5.0           & 5.0   \\
         \midrule
        \textbf{Ratio}     & 0.57        & 0.59        & 0.71       & 0.71       & 0.77         & 0.77         & 0.77          & 0.77   \\
        \bottomrule
    \end{tabular}
\end{adjustbox}
   \caption{Multitasking statistics of synthetic logs for multitasking capacity evaluation.}
   \label{tbl:multi-synt-log-desc}
 \end{table}

 \begin{table}[tp]
    \centering
    \small
\begin{adjustbox}{width=0.65\textwidth}
    \begin{tabular}{{lcccc}}
        \toprule
                                   & {\bf AC-CRE}   & {\bf GOV} & {\bf WK-ORD}  & {\bf PAY}  \\
        \midrule
        \textbf{Mean}              & 1.27        & 5.57          & 5.27    & 817.0         \\
         \midrule
        \textbf{Median}            &  1.0        & 4.0           &  2.0    & 985.0        \\
         \midrule
        \textbf{Ratio}             &  0.14       & 0.86          & 0.64    & 0.99        \\
        \bottomrule
    \end{tabular}
\end{adjustbox}
   \caption{Multitasking statistics of real logs for multitasking capacity evaluation.}
   \label{tbl:multi-real-log-desc}
 \end{table}

The real-life logs cover varying levels of multitasking (see Table~\ref{tbl:multi-real-log-desc}). The first event log {\it AC-CRE} is an anonymized log of an academic recognition process at a university. In this process, a worker generally performs one task at a time, but occasionally, a worker may take on a second or a third activity instance concurrently. This event log contains a large proportion of resources with low participation in the process, meaning that each of these resources performs only a handful of activity instances across the entire period covered by the event log.\footnote{These resources are academic staff who need to approve requests for credentials within their narrow area of expertise.} This explains the large number of resources in this process.

The second event log corresponds to an application-to-approval process in a government agency ({\it GOV}). In this process, each worker handles multiple applications concurrently. The third dataset ({\it WK-ORD}) is an anonymized version of an event log of a customer service process at a utility company. In this process, clients create tickets to report an issue. The tickets give rise to work orders, which workers in the backend oversee. In about a third of the cases, a worker oversees only one work order at a time. In the other two-thirds of cases, a worker oversees multiple work orders, usually related to the same incident.
Finally, the fourth dataset, namely {\it PAY}, is an anonymized version of a payments monitoring process in a financial institution. In this process, each worker monitors hundreds of payments at a time (e.g., a chunk of a company's payroll). The worker performs several cross-checks and moves forward the payments, usually in bulk actions.

\subsubsection{Datasets for Evaluating the Resource Unavailability Impact}\label{sect:data:impact}

The experiments corresponding to Evaluation Question {\bf III} rely on 52 synthetic logs. They were generated from the same loan application process model used on the availability calendars evaluation, all including 2000 process cases and 17 activities. To guarantee different unavailability periods, we modified the following generation steps in the log generation:

\begin{itemize}
    \item From the resource perspective, we created two scenarios: one with a single resource and one with multiple resources. Accordingly, we reduced the number of resources in the original model to one and two resources, respectively. For both scenarios, we assigned a random distribution function from the original resource candidates to generate the processing time for each activity. Both resources were assigned a traditional 8-hour working calendar (Monday-Friday, 08:00-12:00 and 13:00-17:00) assuming a balanced workload.
    \item To avoid an excessive overload affecting the accuracy, we reduced the inter-arrival ratio to align with the reduced number of resources. We also created two scenarios to consider different workloads. Specifically, we generated event logs with 2000 cases and inter-arrival times following a Poisson distribution with a mean of 4 and 8 cases per day, respectively, on the same 8-hour working calendar as the resources.
\end{itemize}

\begin{table}[tp]
    \centering
    \small
\begin{adjustbox}{width=0.99\textwidth}

    \begin{tabular}{{lcccc}}
        \toprule
         & \textbf{Loan$_{R1}^{C4}$ /} & \textbf{Loan$_{R1}^{C8}$ /} & \textbf{Loan$_{R2}^{C4}$ /} & \textbf{Loan$_{R2}^{C8}$ /}  \\
        \midrule
        \textbf{Events}              & 23400 & 23444 & 23276 &  23458   \\
        \midrule
        \textbf{R1 Max Multitask}    & 20    & 39    & 11    &  21      \\
        \midrule
        \textbf{R2 Max Multitask}    & -     & -     & 12    &  21   \\
        \bottomrule
    \end{tabular}
    
\end{adjustbox}
   \caption{General characteristics of event logs for resource unavailability impact evaluation.}
   \label{tbl:log-break}
 \end{table}

Combining the scenarios described above, we produced four base event logs. Table~\ref{tbl:log-break} describes the number of instances and maximum multitasking capacity of each involved resource. The supra-indexes $C4$ and $C8$ describe the inter-arrival ratio considered in the generation, while the sub-indexes $R1$ and $R2$ represent the number of resources involved in the process execution.

Note that the base logs assume resources are continuously working without extended breaks. Thus, they assess our approach under ideal conditions in which resources entirely work when training the models and applied later under the assumption that resources are consistently available according to their calendars. However, three scenarios can affect accuracy when a resource becomes unavailable:
\begin{itemize}
    \item {\sc TRAIN}: Model is trained while the resource is on vacation but later applied when the resource is fully working,
    \item {\sc TEST}: Model is trained while the resource is fully working but later applied when the resource goes on vacation.
    \item {\sc T\&T}: Model is trained and applied when the resource is on vacation.
\end{itemize}

To emulate these scenarios, we modified each base log to reproduce the three cases, each perturbated by adding varying levels of resource unavailability, specifically, 1, 2, 4, and 8 weeks of vacation, respectively. This approach led to 12 extra event logs from each base, totaling 54, to assess the impact of resource unavailability impact.

Specifically, for each base log, we introduced the perturbations at different points in the timeline: starting at 10\% of the sorted datetimes in the base log for the {\sc TRAIN} case, starting at 60\% of the datetimes for the {\sc TEST} case, and at both 10\% and 60\% for the {\sc T\&T} case. This approach ensures that during the evaluation phase, when the log is split into training and testing partitions, the resource will perform some tasks according to their original schedules before going into a period of unavailability.

Additionally, the perturbation method varied between single-resource and multiple-resource settings. For a single-resource scenario, this involved shifting all activities' start and completion times to the right, i.e., adding the duration of the resting interval to each timestamp. Consequently, the resource was entirely unavailable during the break period, and consequently, all activities became delayed from that point. For multiple-resource settings, during the break period when the first resource was unavailable, the second resource took over all tasks being executed by the first resource and started all newly enabled tasks. This reallocation increased the workload and multitasking capacity of the second resource. We kept the same inter-arrival ratios for the base event logs in all the cases, as they are independent of the resource schedules in practice.

\subsection{Experiment Setup and Metrics}

To avoid data leakage and overfitting, we split each log into two sets (training and testing) using a temporal split: the first 50\% of traces in chronological order are put in the training dataset (to discover a simulation model), and the other 50\% is used for measuring the accuracy of the model. We evaluate the accuracy of each simulation model by simulating it using {\sc Prosimos} and measuring the disparity between the simulated output and the testing subset of the log.

Chapela-Campa et al.~\cite{Chapela-CampaBBDKS22} propose multiple metrics to evaluate the quality of data-derived simulation models along the control flow, temporal, and congestion dimensions. Since our study focuses on resource availability and capacity, we only consider temporal and congestion metrics. In the congestion dimension, the authors propose two metrics: arrival times and cycle times. We discard the inter-arrival metric (out of scope). Instead, we take the arrival times from the testing log, so that errors in estimating arrival times do not affect the results. Thus, we only retain the Cycle Time Distribution (CTD) metric. Along the temporal dimension, Chapela-Campa et al. propose three metrics: Circadian Event Distribution (CED), Absolute Event Distribution (AED), and Relative Event Distribution (RED). Since our approach does not estimate inter-arrival times nor seasonality, we omit the CED and AED metrics (as the latter is heavily dependent on the accuracy of the inter-arrival times). Thus, we retain the RED metric, which quantifies the ability of a simulation model to replicate the occurrence of events (and their datetimes) from the beginning of a case to its end.

The CTD metric evaluates the simulation model's capacity to replicate the overall cycle time of a process. It derives empirical Probability Distribution Functions (PDF) from histograms of cycle times collected from two event logs, {\it L1} and {\it L2}, and calculates the CTD distance as the 1-Wasserstein Distance (1WD)~\cite{LevinaB01} between these histograms. The Relative RED analyses the ability of the simulator to mimic the temporal distribution of events relative to the origin of the case. It adjusts all datetimes in {\it L1} and {\it L2} to originate from their respective case arrival times (i.e., the initial datetime in a case is set to 0, with the following datetimes adjusted by the inter-event times). The RED Distance computes the 1WD between the discretized event logs {\it L1} and {\it L2}. Note that both the RED and the CDT are non-normalized metrics. This is due to the fact that the Wasserstein distance is sensitive to the number of data points and the scale. 
Therefore, the absolute values of the CTD and RED metrics do not allow us to compare the performance of a technique across multiple datasets, but they allow us to compare the performance of different techniques on the same dataset.

Additionally to the RED and CTD, we measured the mismatch ratio (MMR) to determine the resource discrepancy between the real and the simulated logs, i.e., $MMR = 1 - |RProf_{simulated} \cap RProf_{real}|  / |RProf_{real}|$. A score of 0 means that both logs contain exactly the same resources, while a score of 1 indicates that the datasets do not contain any common resources. Here, the ``union'' operation refers to the combined set of all unique profiles $RProf$ from both sets being compared. These profiles do not exist explicitly in the real or simulated logs but are discovered from the real log and then replicated in the simulated log. Consequently, some resources in the real log might be removed or combined, producing new profiles that did not exist in the original log. Thus, profiles in the union do not need to be identical but should be functionally similar or equivalent, representing the same roles or capabilities, even if there are minor attribute variations.

\subsection{Experiment Results: Availability Calendar Modeling}

Tables~\ref{tbl:res_syntetic} and~\ref{tbl:res_real} show the evaluation results conducted on synthetic and real logs, respectively. Besides the probabilistic method (labeled as Prob.), we evaluated two variations of the crisp approach. Crisp calendars filter granules, group, and remove resources whose data in the event log does not meet the pre-set parameters for confidence, support, and participation~\cite{Lopez-PintadoD22}. To avoid resource clustering, the variant labeled naive (N-Crisp) corresponds to a configuration of these parameters set to 0. Meanwhile, the full version (C-Crisp) aligns with the optimal hyperparameters, which could lead to resource groupings. As for the C-Crisp and Prob methods, we compare the results corresponding to the optimal hyperparameters that minimize the RED metric. These were determined using 30 iterations (i.e., the default value recommended by the Python library bayes\_opt\footnote{\url{https://github.com/bayesian-optimization/BayesianOptimization}}) of a Bayesian hyper-parameter optimizer~\cite{SnoekLA12}. Additionally, to mitigate the impact of the simulations' stochastic nature on the results, we executed five simulations in each iteration, excluding the lowest and highest values and retaining the mean value of the remaining three. Tables~\ref{tbl:res_syntetic} and~\ref{tbl:res_real} highlight the best approach, i.e., the most favorable (lower) results in each of the 12 logs assessed according to the RED and CTD metrics. 

\begin{table}[tp]
\centering
\setlength{\tabcolsep}{3.0pt}

\begin{tabularx}{\linewidth}{lccccccccc }
\toprule
&& {\bf B-24} & {\bf U-24} & {\bf B-8} & {\bf U-8} & {\bf B-8/4} & {\bf U-8/4} & {\bf B-24/A} & {\bf U-24/A}  \\
    \midrule
\parbox[t]{1.5mm}{\multirow{3}{*}{\rotatebox[origin=c]{90}{RED}}}  
& N-Crisp & 1.98 & 2.39 & 181.71 & 200.76 & 249.74 & 819.42 & 233.61 & 604.76 \\
& C-Crisp & \cellcolor{blue!25}1.42 & \cellcolor{blue!25}1.91 & 166.67 & 158.3  & 205.87 & 731.06 & 197.03 & 452.15 \\
& Prob.   & 7.03 & 7.82 & \cellcolor{blue!25}158.23 & \cellcolor{blue!25}95.58  & \cellcolor{blue!25}69.56  & \cellcolor{blue!25}186.43 & \cellcolor{blue!25}111.18 & \cellcolor{blue!25}361.13 \\
\midrule
\parbox[t]{0.1mm}{\multirow{3}{*}{\rotatebox[origin=c]{90}{CTD}}}  
& N-Crisp & \cellcolor{blue!25}2.27  & 2.63 & 436.21 & 445.28 & 414.48 & 1100.04 & 335.42 & 858.99 \\
& C-Crisp & 2.70  & \cellcolor{blue!25}2.16 & 383.67 & 383.58 & 328.26 & 924.27  & 259.45 & 564.9 \\
& Prob.   & 13.14 & 16.8 & \cellcolor{blue!25}356.56 & \cellcolor{blue!25}164.96 & \cellcolor{blue!25}149.43 & \cellcolor{blue!25}345.05  & \cellcolor{blue!25}201.23 & \cellcolor{blue!25}536.1 \\
\midrule
\parbox[t]{0.1mm}{\multirow{3}{*}{\rotatebox[origin=c]{90}{MMR}}}  
& N-Crisp & 0.0 & 0.0  & 0.0  & 0.0  & 0.0  & 0.0  & 0.0  & 0.0 \\
& C-Crisp & 0.0 & 0.93 & 0.24 & 0.93 & 0.85 & 0.17 & 0.78 & 0.78 \\
& Prob.   & 0.0 & 0.0  & 0.0  & 0.0  & 0.0  & 0.0  & 0.0  & 0.0 \\
\bottomrule
\end{tabularx}
\caption{Evaluation results for availability calendars in synthetic logs.}
\label{tbl:res_syntetic}
\end{table}

Regarding the synthetic evaluation, Table~\ref{tbl:res_syntetic} shows that the probabilistic approach tends to provide more precise results concerning the RED and CTD metrics across most of the tested configurations. More specifically, the probabilistic model overshadows the crisp one in the U-8, B-8/4, and U-8/4 in both RED and CTD metrics. In the U-8, the resources work full-time, but they have uneven workloads, better captured probabilistically than in a crisp way. In the B-8/4 and U-8/4, half of the resources work part-time, leading to lower frequencies that are more challenging to capture with a crisp model. The combination of part-time and uneven workloads in the U-8/4 spotlight the most significant differences, illustrating a scenario in which probabilistic calendars might be more suitable. Although probabilistic models perform better in B-24/A and U-24/A, the differences are less significant. The latest illustrates a limitation of both models, i.e., they are cyclical (weekly) calendars, and here, some resources do not follow a cyclical schedule. Thus, the probabilistic models better capture the different resource frequencies, but an extra dimension to capture seasonal (acyclic) behavior is missing. 

\begin{table}[tp]
\centering 

\setlength{\tabcolsep}{12.5pt}

\begin{tabularx}{\linewidth}{lcccc }
\toprule
&&            {\bf BPIC12}            & {\bf BPIC17}            & {\bf CALL} \\
    \midrule
\parbox[t]{1.5mm}{\multirow{3}{*}{\rotatebox[origin=c]{90}{RED}}}  
& N-Crisp & 213.42                    & 209.86                                        & 0.63 \\
& C-Crisp & 192.64                    & 199.4                     & \cellcolor{blue!25}0.03 \\
& Prob.   & \cellcolor{blue!25}187.28 & \cellcolor{blue!25}162.74                     & 0.83  \\
    \midrule
\parbox[t]{1.5mm}{\multirow{3}{*}{\rotatebox[origin=c]{90}{CTD}}}  
& N-Crisp & 193.01 & 284.56 & 0.97 \\
& C-Crisp & 175.1  & 271.88 & \cellcolor{blue!25}0.04 \\
& Prob.   & \cellcolor{blue!25}170.55 & \cellcolor{blue!25}227.62 & 1.34  \\
\midrule
\parbox[t]{1.5mm}{\multirow{3}{*}{\rotatebox[origin=c]{90}{MMR}}}  
& N-Crisp & 0.05 & 0.01 & 0.12 \\
& C-Crisp & 1.0  & 0.55 & 0.99 \\
& Prob.   & 0.05 & 0.01 & 0.13 \\
\bottomrule
\end{tabularx}
\caption{Evaluation results for availability calendars in real logs.}
\label{tbl:res_real}
\end{table}

In Table~\ref{tbl:res_syntetic}, the N-Crisp approach has superior accuracy only in B-24 and U-24 scenarios, which are highly crisp schedules due to the always available resources (24/7). The probabilistic approach underperforms here because it only assigns a 100\% probability to a time slot if every resource starts an activity instance in every single granule in its calendar every day, which is improbable. The crisp approach, in contrast, assigns a 100\% availability to every granule that achieves the required confidence and support levels. Finally, in 5 out of 8 logs, crisp calendars reached MMR ratios over 0.7, indicating that replicating time accuracy comes at the expense of grouping resources, retaining only 30\% or less. The latest could challenge the (specific) analysis of resource behaviors and the detection of resource-related issues. Conversely, probabilistic calendars retain all original resources while offering comparable or superior time estimation accuracy.

Table~\ref{tbl:res_real} also highlights a superior accuracy of the probabilistic approach in the assessed real-life logs, with the lowest RED in BPIC12 and BPIC17 and the lowest CTD in all the logs, except the CALL log where C-Crisp outperforms. The CALL log is the only dataset where the Crisp calendars outperform the Probabilistic ones among the real-life logs evaluated regarding the RED and CED metrics. Although the Probabilistic model results are close to 1, indicating acceptable accuracy, the Crisp models' lower scores suggest that the events in the CALL log are of a crisp scheduling nature. However, the high MMR values across all the C-Crisp models highlight a trade-off between temporal prediction accuracy and reproducing the exact composition of resources, a limitation absent in the probabilistic and N-Crisp models, which only excluded resources absent in the training datasets.

\subsection{Experiment Results: Multitasking Capacity Modeling}

Tables~\ref{tbl:res_syntetic_multitask} and~\ref{tbl:res_real_multitask} show the evaluation results conducted on synthetic and real logs to assess the multitasking capacity modeling approaches. We evaluated the global ($G_{MULT}$) and local ($L_{MULT}$) multitasking capacity modeling approaches, compared to the probabilistic approach evaluated in the previous section ($S_{PROB}$). Like in the calendar evaluation, we compare the results corresponding to the optimal hyperparameters that minimize the RED metric for the three approaches under assessment. These were determined using 30 iterations. Also, to mitigate the impact of the simulations' stochastic nature on the results, we executed five simulations in each iteration, excluding the lowest and highest values and retaining the mean value of the remaining three. Tables~\ref{tbl:res_syntetic_multitask} and~\ref{tbl:res_real_multitask} highlight the best approach, i.e., the most favorable (lower) results in each of the 12 logs assessed according to the RED and CTD metrics. We omitted the MMR metric in Table~\ref{tbl:res_real_multitask} as in all the synthetic experiments; the models achieved a perfect score of 0.0 for the MMR.

Regarding the synthetic evaluation, Table~\ref{tbl:res_syntetic_multitask} shows that the global approach tends to provide more precise results regarding the RED and CTD metrics in most of the tested configurations, i.e., in 75 \% of the cases. Although the global ($G_{MULT}$) model often outperforms the local ($L_{MULT}$) model, the difference between them is quite narrow. This contrast becomes more evident compared to the single-task ($S_{PROB}$) model's notable underperformance. As expected, the single-task model falls behind in all scenarios due to its inherent limitation of processing enabled activities one at a time, thereby increasing waiting times due to resource contention.

\begin{table}[tp]
\centering
\setlength{\tabcolsep}{2.7pt}

\begin{tabularx}{\linewidth}{lccccccccc }
\toprule
&& {\bf B'-24} & {\bf U'-24} & {\bf B'-8} & {\bf U'-8} & {\bf B'-8/4} & {\bf U'-8/4} & {\bf B'-24/A} & {\bf U'-24/A}  \\
    \midrule
\parbox[t]{1.5mm}{\multirow{3}{*}{\rotatebox[origin=c]{90}{RED}}}  
& $S_{PROB}$ & 142.19 & 150.13 & 317.03 & 410.97 & 605.97 & 629.33 & 382.32 & 360.76 \\
& $G_{MULT}$ & \cellcolor{blue!25}0.94   & \cellcolor{blue!25}1.01   & 1.1    & \cellcolor{blue!25}1.02   & \cellcolor{blue!25}3.03   & \cellcolor{blue!25}2.75   & \cellcolor{blue!25}5.78   & 4.49 \\
& $L_{MULT}$ & \cellcolor{blue!25}0.94   & 1.04   & \cellcolor{blue!25}1.04   & \cellcolor{blue!25}1.02   & 3.07  & 2.76   & 6.01   & \cellcolor{blue!25}4.4 \\
\midrule
\parbox[t]{0.1mm}{\multirow{3}{*}{\rotatebox[origin=c]{90}{CTD}}}  
& $S_{PROB}$ & 230.71 & 248.04 & 605.72 & 790.95 & 1168.7 & 1161.74 & 544.14 & 521.12 \\
& $G_{MULT}$ & \cellcolor{blue!25}1.28   & \cellcolor{blue!25}1.88   & \cellcolor{blue!25}4.43   & 5.41   & \cellcolor{blue!25}10.14  & \cellcolor{blue!25}8.7     & 8.7    & \cellcolor{blue!25}8.98 \\
& $L_{MULT}$ & 1.38   & 1.92   & 5.05   & \cellcolor{blue!25}5.02   & 10.37  & 12.2    & \cellcolor{blue!25}7.54   & 9.57 \\
\bottomrule
\end{tabularx}
\caption{Evaluation results for multitasking capacity in synthetic logs.}
\label{tbl:res_syntetic_multitask}
\end{table}

The first six synthetic scenarios in Table~\ref{tbl:res_syntetic_multitask}, spanning 24-hour full-time, standard 8-hour workdays, and mixed schedules with part-time 4-hour shifts, highlight the following pattern: a decrease in the number of resource working hours (according to their calendars), combined with an increase of the multitasking levels leads to an increase of the values of the RED and CTD metrics. This trend suggests that reducing the number of hours resources are available, combined with higher levels of multitasking, contributes to increased complexity in task management, thereby impacting these performance metrics. In the multitasking models, metric increases are mild, reflecting their ability to manage concurrent tasks despite reduced working hours and higher multitasking levels. In contrast, limited to linear activity processing, the single-task model faces a more significant impact under these conditions, leading to notable RED and CTD metrics increases. 

Despite higher multitasking levels, the decrease in RED and CTD metrics for the single-task model in the B'-24/A and U'-24/A logs in Table~\ref{tbl:res_syntetic_multitask} can be linked to the unique time distribution in these calendars. In these scenarios, resources involved in multitasking might enter long rest periods with multiple ongoing tasks, while in the single-task model, they rest with at most one activity under execution. Additionally, these calendars allow a group of resources to work every day, including weekends, thus increasing total available working hours compared to the part-time schedules in the previous log. The latter leads to lower metric values for the single-task model due to more evenly distributed workloads and greater availability. However, despite this trend, the local and global multitasking models still perform significantly better than the single-task model, effectively managing the activity batches and rest periods inherent in these calendar setups.

The results on real-life logs (Table~\ref{tbl:res_real_multitask}) also highlight a superior accuracy of the multitasking models (both w.r.t., RED and CTD metrics) relative to the single-task model (where multitasking is not allowed). In this evaluation, the global and local multitasking models achieve similar performance. The global model exhibits better results than the local in both metrics on the {\it GOV} and {\it AC-CRE}. In these logs, all approaches show MMR ratios above 0.69, suggesting that only 32\% (or less) of resources in the training log appeared in the testing log, potentially introducing noise. Also, the {\it AC-CRE} log has a low mean multitasking level of 1.27, with approximately 86 \% of events executed independently. The latter allows the single-task model to perform better but is still inferior to both multitasking models. 

\begin{table}[tp]
\centering 

\setlength{\tabcolsep}{11.5pt}
\begin{tabularx}{\linewidth}{lccccc }
\toprule
&& {\bf GOV} & {\bf AC-CRE} & {\bf PAY} & {\bf WK-ORD} \\
    \midrule
\parbox[t]{1.5mm}{\multirow{3}{*}{\rotatebox[origin=c]{90}{RED}}}  
& $S_{PROB}$    & 2148.84 & 72.64 & 178595.39 & 2347.46 \\
& $G_{MULT}$& \cellcolor{blue!25}45.57   & \cellcolor{blue!25}32.9  & 11.53     & 24.65 \\
& $L_{MULT}$ & 45.93   & 36.25 & \cellcolor{blue!25}11.06     & \cellcolor{blue!25}20.36  \\
    \midrule
\parbox[t]{1.5mm}{\multirow{3}{*}{\rotatebox[origin=c]{90}{CTD}}}  
& $S_{PROB}$     & 3155.57 & 100.38 & 295288.06 & 12323.12 \\
& $G_{MULT}$ & \cellcolor{blue!25}29.84   & \cellcolor{blue!25}61.41  & 21.6      & 56.16 \\
& $L_{MULT}$ & 31.65   & 67.35  & \cellcolor{blue!25}20.8      & \cellcolor{blue!25}50.72  \\
\midrule
\parbox[t]{1.5mm}{\multirow{3}{*}{\rotatebox[origin=c]{90}{MMR}}}  
& $S_{PROB}$ & 0.083 & 0.691 & 0 & 0.149 \\
& $G_{MULT}$ & 0.083 & 0.708 & 0 & 0.151 \\
& $L_{MULT}$ & 0.083 & 0.719 & 0 & 0.158 \\
\bottomrule
\end{tabularx}
\caption{Evaluation results for multitasking capacity in real logs.}
\label{tbl:res_real_multitask}
\end{table}

On the other hand, the local model performs better in the {\it PAY} and {\it WK-ORD} logs, as shown in Table~\ref{tbl:res_real_multitask}. These logs are the biggest ones used in the multitasking evaluation regarding the number of process cases and events. As a result, the single-task model achieved the worst scores in the entire evaluation due to the accumulation of enabled activities to be executed in sequence, thus leading to bottlenecks and increasing the waiting times. Conversely, both multitasking models achieved notable scores, showing their capability of handling high multitasking levels. Specifically, the {\it PAY} process had a mean multitasking level greater than 800 split into only nine resources. Still, both multitasking models achieved the lowest scores in both metrics in the entire evaluation over real-life datasets.

In summary, the evaluation reveals that both the local ($L_{MULT}$) and global ($G_{MULT}$) multitasking models consistently outperform the single-task model ($S_{PROB}$) across various settings. The global model, simpler in design, is suitable for broader overviews. In contrast, with its detailed temporal constraints, the local model is better suited for more specific, in-depth, what-if scenario evaluations. Thus, the choice between global or local models relies on the specific needs of the process analysts and the simulation goals.

\subsection{Experiment Results: Resource Unavailability Impact}

To assess the impact of resource unavailability, we used the base logs as a baseline, representing ideal conditions where resources are fully available according to their schedules. We then estimated the accuracy ratio by dividing each perturbed log's metric results (RED and CTD) by the baseline values.

Figure~\ref{fig:red_global} shows the RED ratios, $RED(Perturbation)/ RED(baseline)$, for different scenarios discovered using the global multitasking model. Each subfigure represents a distinct setup based on the number of resources and cases processed daily. Subfigures (a) and (b) represent setups with one resource handling 4 and 8 cases per day, respectively, while (c) and (d) show setups with two resources handling 4 and 8 cases per day. The y-axis shows the RED ratio, and the x-axis groups the scenarios into {\sc TRAIN}, {\sc TEST}, and {\sc T\&T} (see ~Section\ref{sect:data:impact}), with specific values in parentheses indicating the average ratio. Each bar within the subfigures indicates the RED-Ratio for 1-week, 2-week, 4-week, and 8-week break periods, with colors representing different break durations. The RED-Ratio compares the accuracy of the perturbed logs to the baseline: values equal to 1 mean the perturbation achieved the same accuracy as the baseline, values between 0 and 1 indicate better accuracy, and values greater than 1 indicate less accuracy.

\begin{figure}[tp]
    \centering
    \begin{minipage}{0.5\textwidth}
        \centering
        \caption*{{\bf (a) 1 Resource, 4 cases per day}}
        \vspace{-10.0pt}
        \begin{tikzpicture}
            \begin{axis}[
                ybar,
                symbolic x coords={TRAIN, TEST, T\&T},
                xtick=data,
                xtick=data,
                xticklabels={\shortstack{TRAIN\\(1.03)}, \shortstack{TEST\\(1.07)}, \shortstack{T\&T\\(1.06)}},
                ylabel={RED-Ratio},
                enlarge x limits=0.25,
                ymin=0,
                ymax=3.25,
                bar width=7pt,
                width=\textwidth,
                height=0.6\textwidth,
                tick label style={font=\scriptsize}
            ]
            \addplot[ybar, fill=blue, draw=blue] coordinates {(TRAIN,0.91) (TEST,1.007) (T\&T,0.896)};
            \addplot[ybar, fill=orange, draw=orange] coordinates {(TRAIN,0.925) (TEST,1.034) (T\&T,0.937)};
            \addplot[ybar, fill=green, draw=green] coordinates {(TRAIN,1.041) (TEST,1.067) (T\&T,1.071)};
            \addplot[ybar, fill=red, draw=red] coordinates {(TRAIN,1.261) (TEST,1.172) (T\&T,1.351)};
            \end{axis}
        \end{tikzpicture}
    \end{minipage}\hfill
    \begin{minipage}{0.5\textwidth}
        \centering
        \caption*{{\bf (b) 1 Resource, 8 cases per day}}
        \vspace{-10.0pt}
        \begin{tikzpicture}
            \begin{axis}[
                ybar,
                symbolic x coords={TRAIN, TEST, T\&T},
                xtick=data,
                xticklabels={\shortstack{TRAIN\\(0.81)}, \shortstack{TEST\\(1.94)}, \shortstack{T\&T\\(1.35)}},
                ylabel={RED-Ratio},
                enlarge x limits=0.25,
                ymin=0,
                ymax=3.25,
                bar width=7pt,
                width=\textwidth,
                height=0.6\textwidth,
                tick label style={font=\scriptsize}
            ]
            \addplot[ybar, fill=blue, draw=blue] coordinates {(TRAIN,0.822) (TEST,1.241) (T\&T,0.923)};
            \addplot[ybar, fill=orange, draw=orange] coordinates {(TRAIN,0.801) (TEST,1.49) (T\&T,0.986)};
            \addplot[ybar, fill=green, draw=green] coordinates {(TRAIN,0.895) (TEST,2) (T\&T,1.273)};
            \addplot[ybar, fill=red, draw=red] coordinates {(TRAIN,0.731) (TEST,3.024) (T\&T,2.227)};
            \end{axis}
        \end{tikzpicture}
    \end{minipage}

    \centering
    \begin{minipage}{0.5\textwidth}
        \centering
        \caption*{{\bf (c) 2 Resources, 4 cases per day}}
        \vspace{-10.0pt}
        \begin{tikzpicture}
            \begin{axis}[
                ybar,
                symbolic x coords={TRAIN, TEST, T\&T},
                xtick=data,
                xticklabels={\shortstack{TRAIN\\(0.88)}, \shortstack{TEST\\(1.06)}, \shortstack{T\&T\\(0.95)}},
                ylabel={RED-Ratio},
                enlarge x limits=0.25,
                ymin=0,
                ymax=3.25,
                bar width=7pt,
                width=\textwidth,
                height=0.6\textwidth,
                tick label style={font=\scriptsize}
            ]
            \addplot[ybar, fill=blue, draw=blue] coordinates {(TRAIN,0.957) (TEST,1.016) (T\&T,0.973)};
            \addplot[ybar, fill=orange, draw=orange] coordinates {(TRAIN,0.941) (TEST,1.051) (T\&T,0.933)};
            \addplot[ybar, fill=green, draw=green] coordinates {(TRAIN,0.831) (TEST,1.055) (T\&T,0.925)};
            \addplot[ybar, fill=red, draw=red] coordinates {(TRAIN,0.78) (TEST,1.122) (T\&T,0.969)};
            \end{axis}
        \end{tikzpicture}
    \end{minipage}\hfill
    \begin{minipage}{0.5\textwidth}
        \centering
        \caption*{{\bf (d) 2 Resources, 8 cases per day}}
        \vspace{-10.0pt}
        \begin{tikzpicture}
            \begin{axis}[
                ybar,
                symbolic x coords={TRAIN, TEST, T\&T},
                xtick=data,
                xticklabels={\shortstack{TRAIN\\(0.98)}, \shortstack{TEST\\(1.11)}, \shortstack{T\&T\\(1.03)}},
                ylabel={RED-Ratio},
                enlarge x limits=0.25,
                ymin=0,
                ymax=3.25,
                bar width=7pt,
                width=\textwidth,
                height=0.6\textwidth,
                tick label style={font=\scriptsize}
            ]
            \addplot[ybar, fill=blue, draw=blue] coordinates {(TRAIN,0.906) (TEST,1.013) (T\&T,0.919)};
            \addplot[ybar, fill=orange, draw=orange] coordinates {(TRAIN,0.877) (TEST,1.064) (T\&T,0.902)};
            \addplot[ybar, fill=green, draw=green] coordinates {(TRAIN,0.945) (TEST,1.115) (T\&T,1.03)};
            \addplot[ybar, fill=red, draw=red] coordinates {(TRAIN,1.204) (TEST,1.264) (T\&T,1.255)};
            \end{axis}
        \end{tikzpicture}
    \end{minipage}

    \begin{tikzpicture}
        \begin{axis}[
            hide axis,
            xmin=0,
            xmax=1,
            ymin=0,
            ymax=1,
            legend style={at={(0.5,-0.15)}, anchor=north, legend columns=4},
            font=\scriptsize
        ]
        \addlegendimage{area legend, mark=square*, fill=blue, draw=blue}
        \addlegendimage{area legend, mark=square*, fill=orange, draw=orange}
        \addlegendimage{area legend, mark=square*, fill=green, draw=green}
        \addlegendimage{area legend, mark=square*, fill=red, draw=red}
        \legend{1-WEEK BREAK, 2-WEEKS BREAK, 4-WEEKS BREAK, 8-WEEKS BREAK}
        \end{axis}
    \end{tikzpicture}
    \caption{RED: Relative Event Distribution - Global Multitasking Model.}
    \label{fig:red_global}
\end{figure}

We can draw several conclusions based on Figure~\ref{fig:red_global}. As the break duration increases, RED ratios generally increase across all scenarios, indicating that longer unavailability periods lead to decreased accuracy. For shorter breaks (1-week and 2-week), the RED ratios are closer to 1, suggesting that shorter unavailability periods have a minimal impact on accuracy. The RED ratios in the {\sc TRAIN} scenario are generally lower, with accuracy often as good as or better than the baseline. The latter is because training while the resource is on vacation allows the probabilistic approaches to detect unavailable periods and avoid penalizing the probability function.

The {\sc TEST} scenarios show the highest RED ratios, particularly with longer breaks. The increase occurs because the model, trained without resource vacation, is tested during unavailability periods, leading to unobserved behavior and impacting accuracy. The {\sc T\&T} scenarios, which involve training and testing with unavailability periods, show RED Ratios close to 1, indicating comparable accuracy to the baseline, suggesting that the model can adapt to unavailability when considered during training and testing.

In multiple-resource settings, the impact of unavailability is less severe than in single-resource settings, with RED Ratios showing less variability and remaining closer to 1 even in the {\sc TEST} scenarios. Therefore, having multiple resources might help mitigate the adverse effects of unavailability in the time dimension measured by the RED metric.

Figure~\ref{fig:red_local}, assessing the same scenarios as Figure~\ref{fig:red_global} but under the local multitasking model, confirms the observations from the global model. The general trend of increasing RED ratios with longer breaks persists. Still, shorter breaks show minimal impact on accuracy, with RED ratios remaining closer to 1. The {\sc TRAIN} scenarios also display the lowest RED ratios, reinforcing that training during resource unavailability allows the model to adjust. Both models, global and local, show a similar pattern of increased RED ratios in the {\sc TEST} scenarios, with the impact being more noticeable for longer breaks. The {\sc T\&T} scenarios show RED ratios close to 1, indicating the model's ability to adapt when unavailability is considered during training and testing. Regarding accuracy, the global and local models show comparable performance levels in the time dimension assessed by the RED metric across all the scenarios.

\begin{figure}[tp]
    \centering
    \begin{minipage}{0.5\textwidth}
        \centering
        \caption*{{\bf (a) 1 Resource, 4 cases per day}}
        \vspace{-10.0pt}
        \begin{tikzpicture}
            \begin{axis}[
                ybar,
                symbolic x coords={TRAIN, TEST, T\&T},
                ylabel={RED-Ratio},
                xtick=data,
                xticklabels={\shortstack{TRAIN\\(1.02)}, \shortstack{TEST\\(1.11)}, \shortstack{T\&T\\(1.08)}},
                enlarge x limits=0.25,
                ymin=0,
                ymax=3.25,
                bar width=7pt,
                width=\textwidth,
                height=0.6\textwidth,
                tick label style={font=\scriptsize}
            ]
            \addplot[ybar, fill=blue, draw=blue] coordinates {(TRAIN,0.889) (TEST,1.042) (T\&T,0.92)};
            \addplot[ybar, fill=orange, draw=orange] coordinates {(TRAIN,0.946) (TEST,1.069) (T\&T,0.958)};
            \addplot[ybar, fill=green, draw=green] coordinates {(TRAIN,1.119) (TEST,1.126) (T\&T,1.111)};
            \addplot[ybar, fill=red, draw=red] coordinates {(TRAIN,1.115) (TEST,1.211) (T\&T,1.326)};
            \end{axis}
        \end{tikzpicture}
    \end{minipage}\hfill
    \begin{minipage}{0.5\textwidth}
        \centering
        \caption*{{\bf (b) 1 Resource, 8 cases per day}}
        \vspace{-10.0pt}
        \begin{tikzpicture}
            \begin{axis}[
                ybar,
                symbolic x coords={TRAIN, TEST, T\&T},
                xtick=data,
                xticklabels={\shortstack{TRAIN\\(0.84)}, \shortstack{TEST\\(1.98)}, \shortstack{T\&T\\(1.4)}},
                ylabel={RED-Ratio},
                enlarge x limits=0.25,
                ymin=0,
                ymax=3.20,
                bar width=7pt,
                width=\textwidth,
                height=0.6\textwidth,
                tick label style={font=\scriptsize}
            ]
            \addplot[ybar, fill=blue, draw=blue] coordinates {(TRAIN,0.828) (TEST,1.226) (T\&T,0.961)};
            \addplot[ybar, fill=orange, draw=orange] coordinates {(TRAIN,0.81) (TEST,1.538) (T\&T,1.025)};
            \addplot[ybar, fill=green, draw=green] coordinates {(TRAIN,0.9) (TEST,2.057) (T\&T,1.319)};
            \addplot[ybar, fill=red, draw=red] coordinates {(TRAIN,0.81) (TEST,3.104) (T\&T,2.28)};
            \end{axis}
        \end{tikzpicture}
    \end{minipage}

    \centering
    \begin{minipage}{0.5\textwidth}
        \centering
        \caption*{{\bf (c) 2 Resources, 4 cases per day}}
        \vspace{-10.0pt}
        \begin{tikzpicture}
            \begin{axis}[
                ybar,
                symbolic x coords={TRAIN, TEST, T\&T},
                xtick=data,
                xticklabels={\shortstack{TRAIN\\(0.84)}, \shortstack{TEST\\(1.07)}, \shortstack{T\&T\\(0.94)}},
                ylabel={RED-Ratio},
                enlarge x limits=0.25,
                ymin=0,
                ymax=3.25,
                bar width=7pt,
                width=\textwidth,
                height=0.6\textwidth,
                tick label style={font=\scriptsize}
            ]
            \addplot[ybar, fill=blue, draw=blue] coordinates {(TRAIN,0.945) (TEST,1.028) (T\&T,0.992)};
            \addplot[ybar, fill=orange, draw=orange] coordinates {(TRAIN,0.917) (TEST,1.055) (T\&T,0.953)};
            \addplot[ybar, fill=green, draw=green] coordinates {(TRAIN,0.646) (TEST,1.067) (T\&T,0.882)};
            \addplot[ybar, fill=red, draw=red] coordinates {(TRAIN,0.862) (TEST,1.126) (T\&T,0.917)};
            \end{axis}
        \end{tikzpicture}
    \end{minipage}\hfill
    \begin{minipage}{0.5\textwidth}
        \centering
        \caption*{{\bf (d) 2 Resources, 8 cases per day}}
        \vspace{-10.0pt}
        \begin{tikzpicture}
            \begin{axis}[
                ybar,
                symbolic x coords={TRAIN, TEST, T\&T},
                xtick=data,
                xticklabels={\shortstack{TRAIN\\(0.98)}, \shortstack{TEST\\(1.14)}, \shortstack{T\&T\\(0.98)}},
                ylabel={RED-Ratio},
                enlarge x limits=0.25,
                ymin=0,
                ymax=3.25,
                bar width=7pt,
                width=\textwidth,
                height=0.6\textwidth,
                tick label style={font=\scriptsize}
            ]
            \addplot[ybar, fill=blue, draw=blue] coordinates {(TRAIN,0.873) (TEST,1.048) (T\&T,0.952)};
            \addplot[ybar, fill=orange, draw=orange] coordinates {(TRAIN,0.93) (TEST,1.079) (T\&T,0.926)};
            \addplot[ybar, fill=green, draw=green] coordinates {(TRAIN,0.983) (TEST,1.131) (T\&T,0.943)};
            \addplot[ybar, fill=red, draw=red] coordinates {(TRAIN,1.14) (TEST,1.297) (T\&T,1.118)};
            \end{axis}
        \end{tikzpicture}
    \end{minipage}

    \begin{tikzpicture}
        \begin{axis}[
            hide axis,
            xmin=0,
            xmax=1,
            ymin=0,
            ymax=1,
            legend style={at={(0.5,-0.15)}, anchor=north, legend columns=4},
            font=\scriptsize
        ]
        \addlegendimage{area legend, mark=square*, fill=blue, draw=blue}
        \addlegendimage{area legend, mark=square*, fill=orange, draw=orange}
        \addlegendimage{area legend, mark=square*, fill=green, draw=green}
        \addlegendimage{area legend, mark=square*, fill=red, draw=red}
        \legend{1-WEEK BREAK, 2-WEEKS BREAK, 4-WEEKS BREAK, 8-WEEKS BREAK}
        \end{axis}
    \end{tikzpicture}
    \caption{RED: Relative Event Distribution - Local Multitasking Model.}
    \label{fig:red_local}
\end{figure}

Figures~\ref{fig:ctd_global} and~\ref{fig:ctd_local} display the accuracy of the congestion dimension according to the CTD ratio ($CTD(perturbation) / CTD(baseline)$) obtained by the global and local multitasking models, respectively. They confirm several trends observed in the RED ratios. Both models show comparable performance levels in the congestion dimension assessed by the CTD metric across all the scenarios. The {\sc TRAIN} scenarios generally show low CTD ratios, i.e., comparable to the baseline, indicating that training with unavailability periods allows the approach to handle breaks in the event logs when discovering the simulation models. On the other hand, CTD ratios increase with longer break durations, particularly in the {\sc TEST} and {\sc T\&T}scenarios, indicating an increase in waiting times due to resource contention during extended unavailability periods. Conversely to RED ratios results, the impact of unavailability in multiple-resource settings is more significant than in single-resource settings, with higher CTD ratios even in the {\sc TEST} and {\sc T\&T} scenarios.

\begin{figure}[tp]
    \centering
    \begin{minipage}{0.5\textwidth}
        \centering
        \caption*{{\bf (a) 1 Resource, 4 cases per day}}
        \vspace{-10.0pt}
        \begin{tikzpicture}
            \begin{axis}[
                ybar,
                symbolic x coords={TRAIN, TEST, T\&T},
                xtick=data,
                xtick=data,
                xticklabels={\shortstack{TRAIN\\(1.03)}, \shortstack{TEST\\(1.13)}, \shortstack{T\&T\\(1.09)}},
                ylabel={CTD-Ratio},
                enlarge x limits=0.25,
                ymin=0,
                ymax=6,
                bar width=7pt,
                width=\textwidth,
                height=0.6\textwidth,
                tick label style={font=\scriptsize}
            ]
            \addplot[ybar, fill=blue, draw=blue] coordinates {(TRAIN,0.921) (TEST,1.015) (T\&T,0.921)};
            \addplot[ybar, fill=orange, draw=orange] coordinates {(TRAIN,0.911) (TEST,1.057) (T\&T,0.941)};
            \addplot[ybar, fill=green, draw=green] coordinates {(TRAIN,1.037) (TEST,1.165) (T\&T,1.067)};
            \addplot[ybar, fill=red, draw=red] coordinates {(TRAIN,1.261) (TEST,1.293) (T\&T,1.438)};
            \end{axis}
        \end{tikzpicture}
    \end{minipage}\hfill
    \begin{minipage}{0.5\textwidth}
        \centering
        \caption*{{\bf (b) 1 Resource, 8 cases per day}}
        \vspace{-10.0pt}
        \begin{tikzpicture}
            \begin{axis}[
                ybar,
                symbolic x coords={TRAIN, TEST, T\&T},
                xtick=data,
                xticklabels={\shortstack{TRAIN\\(0.96)}, \shortstack{TEST\\(1.91)}, \shortstack{T\&T\\(1.46)}},
                ylabel={CTD-Ratio},
                enlarge x limits=0.25,
                ymin=0,
                ymax=6,
                bar width=7pt,
                width=\textwidth,
                height=0.6\textwidth,
                tick label style={font=\scriptsize}
            ]
            \addplot[ybar, fill=blue, draw=blue] coordinates {(TRAIN,0.883) (TEST,1.232) (T\&T,1.034)};
            \addplot[ybar, fill=orange, draw=orange] coordinates {(TRAIN,0.93) (TEST,1.482) (T\&T,1.066)};
            \addplot[ybar, fill=green, draw=green] coordinates {(TRAIN,1.062) (TEST,1.934) (T\&T,1.39)};
            \addplot[ybar, fill=red, draw=red] coordinates {(TRAIN,0.959) (TEST,2.987) (T\&T,2.339)};
            \end{axis}
        \end{tikzpicture}
    \end{minipage}

    \centering
    \begin{minipage}{0.5\textwidth}
        \centering
        \caption*{{\bf (c) 2 Resources, 4 cases per day}}
        \vspace{-10.0pt}
        \begin{tikzpicture}
            \begin{axis}[
                ybar,
                symbolic x coords={TRAIN, TEST, T\&T},
                xtick=data,
                xticklabels={\shortstack{TRAIN\\(0.86)}, \shortstack{TEST\\(1.91)}, \shortstack{T\&T\\(1.68)}},
                ylabel={CTD-Ratio},
                enlarge x limits=0.25,
                ymin=0,
                ymax=6,
                bar width=7pt,
                width=\textwidth,
                height=0.6\textwidth,
                tick label style={font=\scriptsize}
            ]
            \addplot[ybar, fill=blue, draw=blue] coordinates {(TRAIN,0.997) (TEST,1.224) (T\&T,1.023)};
            \addplot[ybar, fill=orange, draw=orange] coordinates {(TRAIN,0.95) (TEST,1.48) (T\&T,1.221)};
            \addplot[ybar, fill=green, draw=green] coordinates {(TRAIN,0.746) (TEST,1.965) (T\&T,1.764)};
            \addplot[ybar, fill=red, draw=red] coordinates {(TRAIN,0.739) (TEST,2.985) (T\&T,2.731)};
            \end{axis}
        \end{tikzpicture}
    \end{minipage}\hfill
    \begin{minipage}{0.5\textwidth}
        \centering
        \caption*{{\bf (d) 2 Resources, 8 cases per day}}
        \vspace{-10.0pt}
        \begin{tikzpicture}
            \begin{axis}[
                ybar,
                symbolic x coords={TRAIN, TEST, T\&T},
                xtick=data,
                xticklabels={\shortstack{TRAIN\\(1.09)}, \shortstack{TEST\\(2.93)}, \shortstack{T\&T\\(2.32)}},
                ylabel={CTD-Ratio},
                enlarge x limits=0.25,
                ymin=0,
                ymax=6.0,
                bar width=7pt,
                width=\textwidth,
                height=0.6\textwidth,
                tick label style={font=\scriptsize}
            ]
            \addplot[ybar, fill=blue, draw=blue] coordinates {(TRAIN,0.917) (TEST,1.455) (T\&T,1.305)};
            \addplot[ybar, fill=orange, draw=orange] coordinates {(TRAIN,0.791) (TEST,2.003) (T\&T,1.487)};
            \addplot[ybar, fill=green, draw=green] coordinates {(TRAIN,1.008) (TEST,3.037) (T\&T,2.356)};
            \addplot[ybar, fill=red, draw=red] coordinates {(TRAIN,1.652) (TEST,5.233) (T\&T,4.147)};
            \end{axis}
        \end{tikzpicture}
    \end{minipage}

    \begin{tikzpicture}
        \begin{axis}[
            hide axis,
            xmin=0,
            xmax=1,
            ymin=0,
            ymax=1,
            legend style={at={(0.5,-0.15)}, anchor=north, legend columns=4},
            font=\scriptsize
        ]
        \addlegendimage{area legend, mark=square*, fill=blue, draw=blue}
        \addlegendimage{area legend, mark=square*, fill=orange, draw=orange}
        \addlegendimage{area legend, mark=square*, fill=green, draw=green}
        \addlegendimage{area legend, mark=square*, fill=red, draw=red}
        \legend{1-WEEK BREAK, 2-WEEKS BREAK, 4-WEEKS BREAK, 8-WEEKS BREAK}
        \end{axis}
    \end{tikzpicture}

    \caption{CTD: Cycle Time Distribution - Global Multitasking Model.}
    \label{fig:ctd_global}
\end{figure}

Overall, our findings highlight the capability of the discovery approach to handle resource unavailability. However, our model assumes that resources are intermittently available during the intervals captured by the discovered periodic pattern. It does not capture periods where a resource is entirely unavailable during an interval outside the periodic pattern, such as a vacation break that is either aperiodic or periodic, with periodicity unobserved in the training set. Intuitively, the impact of such a vacation break depends on whether all resources who can perform a task take a break simultaneously (no replacement) or only some of the resources take a break, and others remain available to take their workload (with replacement). In the ``no replacement' scenario, the impact is more evident in the time dimension, as the RED metric is affected by the resulting shift in the event timelines. Specifically, the RED ratios increase during the vacation breaks. In the ``with replacement'' scenario, the impact is more significant in the congestion dimension, as the CTD metric shows. Indeed, when one resource goes on vacation, other resources can take over their tasks, leading to higher workloads and increased congestion. The extent of this impact might vary depending on the number of available resources and their multitasking capabilities.

\begin{figure}[tp]
    \centering
    \begin{minipage}{0.5\textwidth}
        \centering
        \caption*{{\bf (a) 1 Resource, 4 cases per day}}
        \vspace{-10.0pt}
        \begin{tikzpicture}
            \begin{axis}[
                ybar,
                symbolic x coords={TRAIN, TEST, T\&T},
                xtick=data,
                xtick=data,
                xticklabels={\shortstack{TRAIN\\(1.04)}, \shortstack{TEST\\(1.12)}, \shortstack{T\&T\\(1.11)}},
                ylabel={CTD-Ratio},
                enlarge x limits=0.25,
                ymin=0,
                ymax=6,
                bar width=7pt,
                width=\textwidth,
                height=0.6\textwidth,
                tick label style={font=\scriptsize}
            ]
            \addplot[ybar, fill=blue, draw=blue] coordinates {(TRAIN,0.895) (TEST,1.038) (T\&T,0.905)};
            \addplot[ybar, fill=orange, draw=orange] coordinates {(TRAIN,0.931) (TEST,1.128) (T\&T,0.957)};
            \addplot[ybar, fill=green, draw=green] coordinates {(TRAIN,1.079) (TEST,1.205) (T\&T,1.151)};
            \addplot[ybar, fill=red, draw=red] coordinates {(TRAIN,1.243) (TEST,1.417) (T\&T,1.422)};
            \end{axis}
        \end{tikzpicture}
    \end{minipage}\hfill
    \begin{minipage}{0.5\textwidth}
        \centering
        \caption*{{\bf (b) 1 Resource, 8 cases per day}}
        \vspace{-10.0pt}
        \begin{tikzpicture}
            \begin{axis}[
                ybar,
                symbolic x coords={TRAIN, TEST, T\&T},
                xtick=data,
                xticklabels={\shortstack{TRAIN\\(0.96)}, \shortstack{TEST\\(1.92)}, \shortstack{T\&T\\(1.48)}},
                ylabel={CTD-Ratio},
                enlarge x limits=0.25,
                ymin=0,
                ymax=6,
                bar width=7pt,
                width=\textwidth,
                height=0.6\textwidth,
                tick label style={font=\scriptsize}
            ]
            \addplot[ybar, fill=blue, draw=blue] coordinates {(TRAIN,0.891) (TEST,1.196) (T\&T,1.045)};
            \addplot[ybar, fill=orange, draw=orange] coordinates {(TRAIN,0.904) (TEST,1.471) (T\&T,1.079)};
            \addplot[ybar, fill=green, draw=green] coordinates {(TRAIN,1.002) (TEST,1.994) (T\&T,1.422)};
            \addplot[ybar, fill=red, draw=red] coordinates {(TRAIN,1.06) (TEST,3.004) (T\&T,2.377)};
            \end{axis}
        \end{tikzpicture}
    \end{minipage}

    \centering
    \begin{minipage}{0.5\textwidth}
        \centering
        \caption*{{\bf (c) 2 Resources, 4 cases per day}}
        \vspace{-10.0pt}
        \begin{tikzpicture}
            \begin{axis}[
                ybar,
                symbolic x coords={TRAIN, TEST, T\&T},
                xtick=data,
                xticklabels={\shortstack{TRAIN\\(0.83)}, \shortstack{TEST\\(1.87)}, \shortstack{T\&T\\(1.68)}},
                ylabel={CTD-Ratio},
                enlarge x limits=0.25,
                ymin=0,
                ymax=6,
                bar width=7pt,
                width=\textwidth,
                height=0.6\textwidth,
                tick label style={font=\scriptsize}
            ]
            \addplot[ybar, fill=blue, draw=blue] coordinates {(TRAIN,0.968) (TEST,1.197) (T\&T,1.167)};
            \addplot[ybar, fill=orange, draw=orange] coordinates {(TRAIN,0.948) (TEST,1.453) (T\&T,1.392)};
            \addplot[ybar, fill=green, draw=green] coordinates {(TRAIN,0.579) (TEST,1.936) (T\&T,1.7)};
            \addplot[ybar, fill=red, draw=red] coordinates {(TRAIN,0.818) (TEST,2.911) (T\&T,2.458)};
            \end{axis}
        \end{tikzpicture}
    \end{minipage}\hfill
    \begin{minipage}{0.5\textwidth}
        \centering
        \caption*{{\bf (d) 2 Resources, 8 cases per day}}
        \vspace{-10.0pt}
        \begin{tikzpicture}
            \begin{axis}[
                ybar,
                symbolic x coords={TRAIN, TEST, T\&T},
                xtick=data,
                xticklabels={\shortstack{TRAIN\\(1.04)}, \shortstack{TEST\\(2.92)}, \shortstack{T\&T\\(2.47)}},
                ylabel={CTD-Ratio},
                enlarge x limits=0.25,
                ymin=0,
                ymax=6.0,
                bar width=7pt,
                width=\textwidth,
                height=0.6\textwidth,
                tick label style={font=\scriptsize}
            ]
            \addplot[ybar, fill=blue, draw=blue] coordinates {(TRAIN,0.878) (TEST,1.454) (T\&T,1.289)};
            \addplot[ybar, fill=orange, draw=orange] coordinates {(TRAIN,0.865) (TEST,1.968) (T\&T,1.557)};
            \addplot[ybar, fill=green, draw=green] coordinates {(TRAIN,0.926) (TEST,3.032) (T\&T,2.472)};
            \addplot[ybar, fill=red, draw=red] coordinates {(TRAIN,1.488) (TEST,5.215) (T\&T,4.544)};
            \end{axis}
        \end{tikzpicture}
    \end{minipage}

    \begin{tikzpicture}
        \begin{axis}[
            hide axis,
            xmin=0,
            xmax=1,
            ymin=0,
            ymax=1,
            legend style={at={(0.5,-0.15)}, anchor=north, legend columns=4},
            font=\scriptsize
        ]
        \addlegendimage{area legend, mark=square*, fill=blue, draw=blue}
        \addlegendimage{area legend, mark=square*, fill=orange, draw=orange}
        \addlegendimage{area legend, mark=square*, fill=green, draw=green}
        \addlegendimage{area legend, mark=square*, fill=red, draw=red}
        \legend{1-WEEK BREAK, 2-WEEKS BREAK, 4-WEEKS BREAK, 8-WEEKS BREAK}
        \end{axis}
    \end{tikzpicture}

    \caption{CTD: Cycle Time Distribution - Local Multitasking Model.}
    \label{fig:ctd_local}
\end{figure}

In summary, the designed simulation model should be applied to scenarios where resources work intermittently but without going on extended vacations. Explicitly handling vacation breaks in the simulation model is beyond the scope of this paper, constituting a limitation to be addressed in future works.

\subsection{Threats to validity}

The findings are subject to the following threats: (1) \textit{internal validity}: the experiments rely on 68 synthetic and 7 real-life logs. The results could differ on other logs. We selected logs with varying characteristics and from different domains to mitigate this limitation. (2) \textit{ecological validity}: we compare the simulation outputs against the original log. This allows us to measure how well the simulation models replicate the as-is process, but it does not allow us to assess the accuracy improvements of using probabilistic calendars in a what-if setting, i.e., predicting performance after a change.

\section{Conclusion}
\label{sect:conclusion}

This article presented a business process simulation approach that models intermittent resource availability and multitasking behavior probabilistically. 
Instead of assuming that a resource is either available or not available during a time granule, the proposed approach assumes that a resource may start an activity instance, during a time granule, with a given probability. Similarly, to capture multitasking behavior, the approach assumes that a resource may take on an additional activity instance, given that they are already performing N activity instances, with a certain probability that depends on N and on the current time slot. The article additionally put forward an approach to discovering probabilistic availability calendars and multitasking capacities from an event log of a business process. 

The experimental evaluation indicates that, in the context of BP simulation models discovered from event logs, the use of probabilistic availability calendars leads to simulation models that are more accurate than simulation models with crisp availability calendars.
On the other hand, the evaluation of the multitasking capacity modeling approach shows that an approach where the multitasking probabilities are fixed across time (the so-called global multitasking modeling approach) sometimes outperforms, but other times underperform, models where the multitasking probabilities may vary over time (the local multitasking modeling approach). 
It appears that the relative performance of these approaches is dependent on the dataset. 

The latter observation suggests that further research is required to shed light into the factors that affect the multitasking capacity of resources. It is possible that the multitasking capacity is not (only) dependent on the time of the day or of the week, but also on other factors, such as the types of tasks that a resource is currently performing. For example, a resource might be prone to multitasking when they are performing an activity of type A (e.g., checking the completeness of loan applications), but they are less likely to multitask when performing a more complex activity (e.g., assessing the credit-worthiness of a loan applicant).



Another potential limitation of the proposal is that it assumes that the periodicity of time slots in the availability calendars abides to circadian cycles (i.e., weekly, daily, and hourly periodicity). In practical scenarios, the availability of a resource might fluctuate along circannual (seasonal) cycles, e.g., availability differing between summer and winter or even within monthly cycles (e.g.,  availability in January is different from August). Another direction for future research is to extend the proposed approach to capture seasonal and non-periodical availability patterns.



\medskip\noindent\textbf{Reproducibility.} The source code, datasets, models, and instructions to reproduce the experiments can be found at: \url{https://github.com/orlenyslp/probabilistic_resource_calendars}.

\medskip
\noindent\textbf{Acknowledgments.}
This work was funded by the European Research Council (PIX project)

\bibliographystyle{model1-num-names}
\bibliography{references}

\end{document}